\PassOptionsToPackage{dvipsnames}{xcolor}
\PassOptionsToPackage{hyperfootnotes=false}{hyperref}

\documentclass[11pt]{article}
\usepackage[preprint]{acl}

\usepackage{times}
\usepackage{latexsym}
\usepackage[T1]{fontenc}
\usepackage[utf8]{inputenc}
\usepackage{microtype}
\usepackage{inconsolata}
\usepackage{graphicx}

\usepackage{multirow}
\usepackage{booktabs}
\usepackage{subcaption}
\usepackage{hhline}
\usepackage{amsmath}
\usepackage{wasysym}
\usepackage{pifont}
\usepackage{algorithm}
\usepackage{algpseudocode}
\usepackage{amssymb}
\usepackage{enumitem}
\usepackage{adjustbox}
\usepackage{caption}
\usepackage[skins,breakable]{tcolorbox}
\usepackage{color, colortbl}
\usepackage{xcolor}
\usepackage{bbm}
\usepackage{tabularx}
\usepackage{array}
\usepackage{fontawesome5}
\usepackage{threeparttable}
\usepackage{float}
\usepackage{balance}

\definecolor{RedOrange}{RGB}{255,69,0}
\definecolor{myup}{HTML}{A52A2A}
\definecolor{deepblue}{HTML}{1E3A8A}
\definecolor{softpurple}{HTML}{7C3AED}
\definecolor{darkgreen}{HTML}{065F46}
\definecolor{lightgray}{HTML}{F9FAFB}
\definecolor{maincolor}{RGB}{0, 102, 102}
\definecolor{backcolor}{RGB}{245, 250, 250}
\definecolor{forestgreen}{RGB}{34, 139, 34}
\definecolor{blue}{HTML}{C5D3E3}
\definecolor{rowcolor}{HTML}{D4FAFC}

\newcommand{\paratitle}[1]{\vspace{0.8ex}\noindent\textbf{#1}}
\renewcommand{\cite}{\citep}

\graphicspath{{figures/}}

\title{PRISM: A Personalized Dual-Memory Self-Evolving Framework}

\author{%
  Qijia Zhuang\textsuperscript{1},
  Xueming Han\textsuperscript{2},
  Jiaheng Liu\textsuperscript{1},
  Yuyao Zhang\textsuperscript{2}, \\
  \bfseries Fanyu Meng\textsuperscript{2},
  Junlan Feng\textsuperscript{2},
  Yuxiang Ren\textsuperscript{1}\thanks{Corresponding author: Yuxiang Ren.} \\
  \normalfont\textsuperscript{1}\,Nanjing University
  \quad
  \textsuperscript{2}\,Jiutian Research
}
\hypersetup{
  pdftitle={PRISM: A Personalized Dual-Memory Self-Evolving Framework},
  pdfauthor={Qijia Zhuang, Xueming Han, Jiaheng Liu, Yuyao Zhang, Fanyu Meng, Junlan Feng, Yuxiang Ren}
}

\begin{document}
\setcounter{footnote}{1}
\maketitle

\begin{abstract}
Large language model (LLM)-based agents increasingly rely on external memory to support long-horizon, user-centric interactions. However, most existing memory frameworks treat memory as a static store-and-retrieve component, limiting their ability to capture evolving user needs and generate personalized responses. We propose \textbf{PRISM}, a personalized dual-memory self-evolving framework that jointly models \emph{user memory} and \emph{agent memory}. PRISM stores user-related information in a structured memory graph and learns task-oriented strategies from long-term interaction trajectories and feedback signals. It further introduces a self-evolution process that refines the interaction pool and optimizes retrieval and response strategies over time. Experiments on Mem-PAL and KnowMeBench show that PRISM consistently outperforms representative memory-augmented baselines, improving long-term memory recall, personalized reasoning, and user-aligned response generation.
\end{abstract}

\section{Introduction}

In long-horizon user-centric agent tasks, Memory-Augmented Agents have become an important solution~\cite{jiang2024long,wu2025human,fang2025lightmem}. 
They typically rely on external memory systems to store long-term information~\cite{laban2025llms,zhou2025mem1} and retrieve it on demand for decision-making~\cite{wei2022chain}. 
Existing studies mainly maintain user-specific details, such as preferences and identity profiles, to support personalized and contextually consistent interactions~\cite{liu2025survey,mem0,zhong2024memorybank,kang2025memory}. 
Beyond static memory storage, recent self-evolving memory frameworks further improve agents through experience accumulation, reflection-based trajectory abstraction~\cite{liang2025sage,cao2025remember}, or training-free policy optimization via semantic comparisons~\cite{cai2025training}.

However, in user-centric agent settings, current memory frameworks still face several fundamental limitations.

\paratitle{Rigid Strategies and Retrieval Mechanisms.}
Personalized responses require not only accurate memory retrieval, but also retrieval and response strategies adapted to the target user~\cite{zeng2024personalized}. 
For example, explaining ``love'' may require a picture book for a child but a song for a music enthusiast, showing that retrieval content and response style should vary across users~\cite{zhang2025personalize}. 
However, existing systems often treat user profiling and memory retrieval as independent processes, leading to fixed decision patterns that fail to capture subtle user differences~\cite{wu2024understanding}. 
Moreover, flat memory architectures struggle to distinguish temporal, semantic, and entity-related facets, often introducing redundant or noisy information during retrieval~\cite{liu2024lost,gutierrez2025rag}.

\paratitle{Ignoring Personalized Optimization Objectives.}
Most existing self-evolving memory frameworks optimize task-level success on static datasets, while overlooking user alignment~\cite{cai2026one}. 
However, personalized agents require not only correct task completion, but also alignment with individual preferences and constraints. 
Without explicit feedback-based optimization~\cite{guan2025survey}, agents may satisfy general task goals while deviating from users' long-term needs. 
In contrast, feedback-driven reinforcement learning provides user-specific signals for continuously adapting memory retrieval and response strategies.

\begin{figure}[t] 
    \centering
    \resizebox{1\columnwidth}{!}{
    \includegraphics{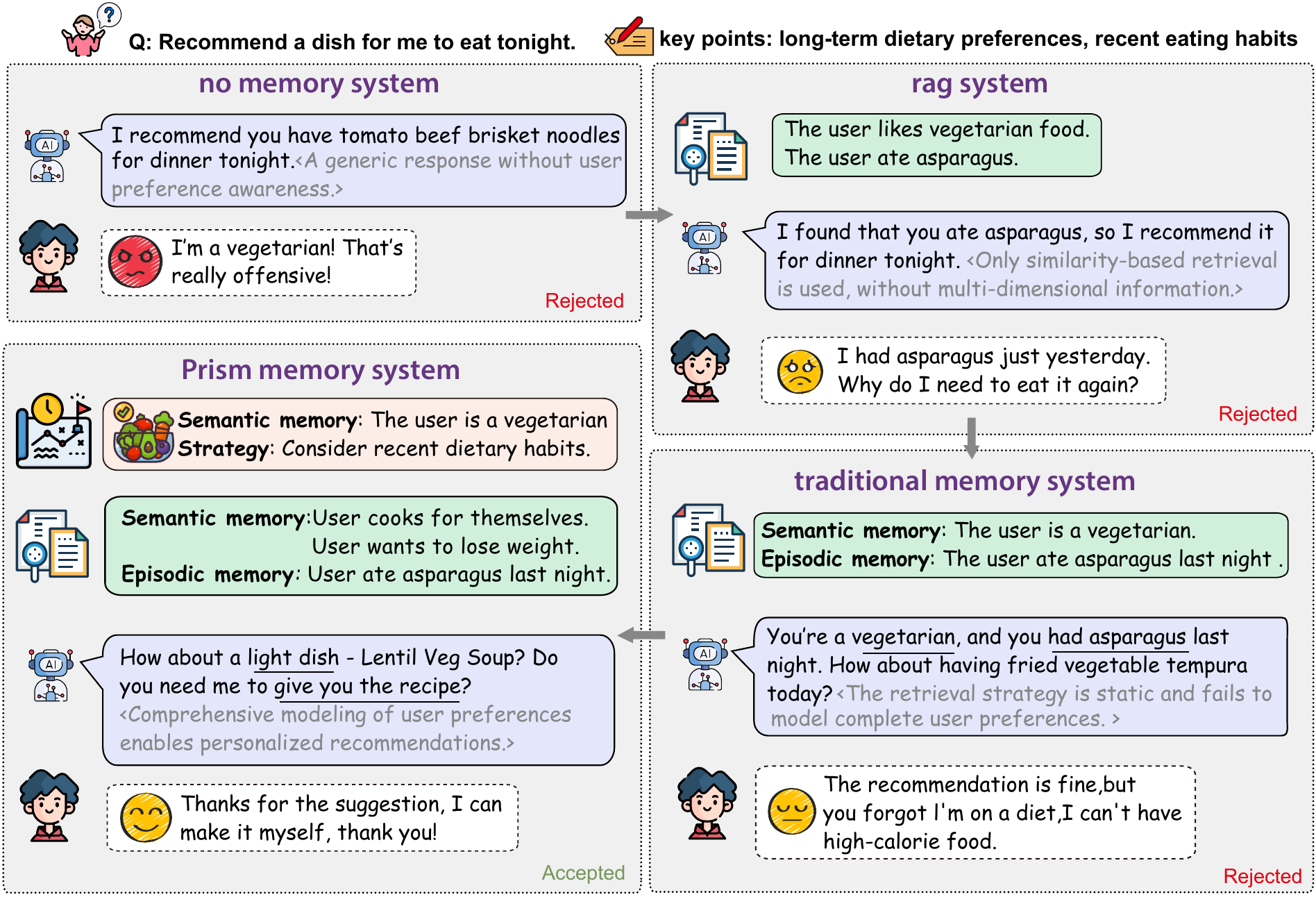}}
    \caption{A comparative illustration of agentic behavior under different memory architectures.}
    \label{fig:subgraph}
    \vspace{-7mm}
\end{figure}

To address these limitations, we propose \textbf{PRISM}, a \textbf{P}e\textbf{r}sonal\textbf{i}zed Dual-Memory \textbf{S}elf-evolving fra\textbf{m}ework that jointly models \emph{user memory} and \emph{agent memory}. 
Instead of merely storing user semantic or episodic information, PRISM learns task-oriented strategies from user feedback and stores them in agent memory. 
Given a new query, the agent retrieves relevant strategies to guide user memory retrieval and response generation, as illustrated in Figure~\ref{fig:subgraph}. 
PRISM further adopts feedback-driven reinforcement learning with two reward components: a task accuracy reward and a user persona reward.

Based on the above design, our main contributions are summarized as follows:
\begin{itemize}
    \item We propose \textbf{PRISM}, a personalized dual-memory framework that jointly models user memory and agent memory, where agent memory provides strategy-level guidance for memory retrieval and response generation.
    \item We introduce a feedback-driven self-evolution mechanism that continuously optimizes the interaction pool and updates the on-device LLM through reinforcement learning, enabling the agent to adapt to evolving user preferences and task requirements.
    \item We conduct extensive experiments on Mem-PAL~\cite{huang2025mem} and KnowMeBench~\cite{wu2026knowme}, demonstrating that PRISM significantly outperforms existing memory-based baselines on complex memory and personalization tasks.
\end{itemize}

\section{Related Works}

\subsection{Agent Memory Systems}
Fixed-length context windows limit LLMs' ability to retain information across long-term interactions, making persistent memory essential for long-horizon agents~\cite{liu2024lost,zhang2025survey,hu2025memory}.
Recent studies augment LLM-based agents with external memory modules for selective storage, retrieval, and updating~\cite{li2025memos,packer2023memgpt}.
Representative systems include Mem0, which introduces a memory CRUD pipeline~\cite{mem0}, and A-Mem, which organizes memory fragments into linked-list structures for efficient retrieval~\cite{xu2025mem}.

Reinforcement learning has also been applied to memory storage, management, and retrieval.
For example, MemoryR1 trains agents on LoCoMo to decide when to update or query external memory~\cite{yan2025memory,maharana2024evaluating}, while MemAlpha trains agents to invoke explicit memory operation tools in multi-turn interactions~\cite{wang2025mem}.

In contrast, PRISM organizes memory into a structured dual-memory architecture and leverages long-term interaction histories for feedback-driven memory evolution and on-device LLM optimization.

\subsection{Personalized Agent}
Personalized modeling for LLM-based agents aims to tailor responses according to user profiles, preferences, and contextual information~\cite{zhang2024personalization}.
Existing methods generally fall into two categories~\cite{salemi2025comparing}: parameterized methods, which inject user information through fine-tuning~\cite{bini2025memlora} or reinforcement learning~\cite{salemi2024optimization,tan2024democratizing,zhu2025towards}, and retrieval-based methods, such as Retrieval-Augmented Generation (RAG) and memory retrieval~\cite{wang2024ai,qiu2025measuring,pan2025memory}.
However, both paradigms have limitations~\cite{li2025personalized}.
Fine-tuning-based approaches require substantial computational cost for each update and struggle to adapt quickly when user preferences change, while direct retrieval-based methods are often constrained by temporal and semantic gaps between concrete behaviors and abstract user profiles.
PRISM addresses these limitations through the continuous evolution of dual memory and iterative model updates, enabling personalized responses that better adapt to dynamic user preferences.

\section{Methodology}
\subsection{Framework Overview}
\begin{figure*}[t]
    \centering
    \resizebox{1\linewidth}{!}{
    \includegraphics{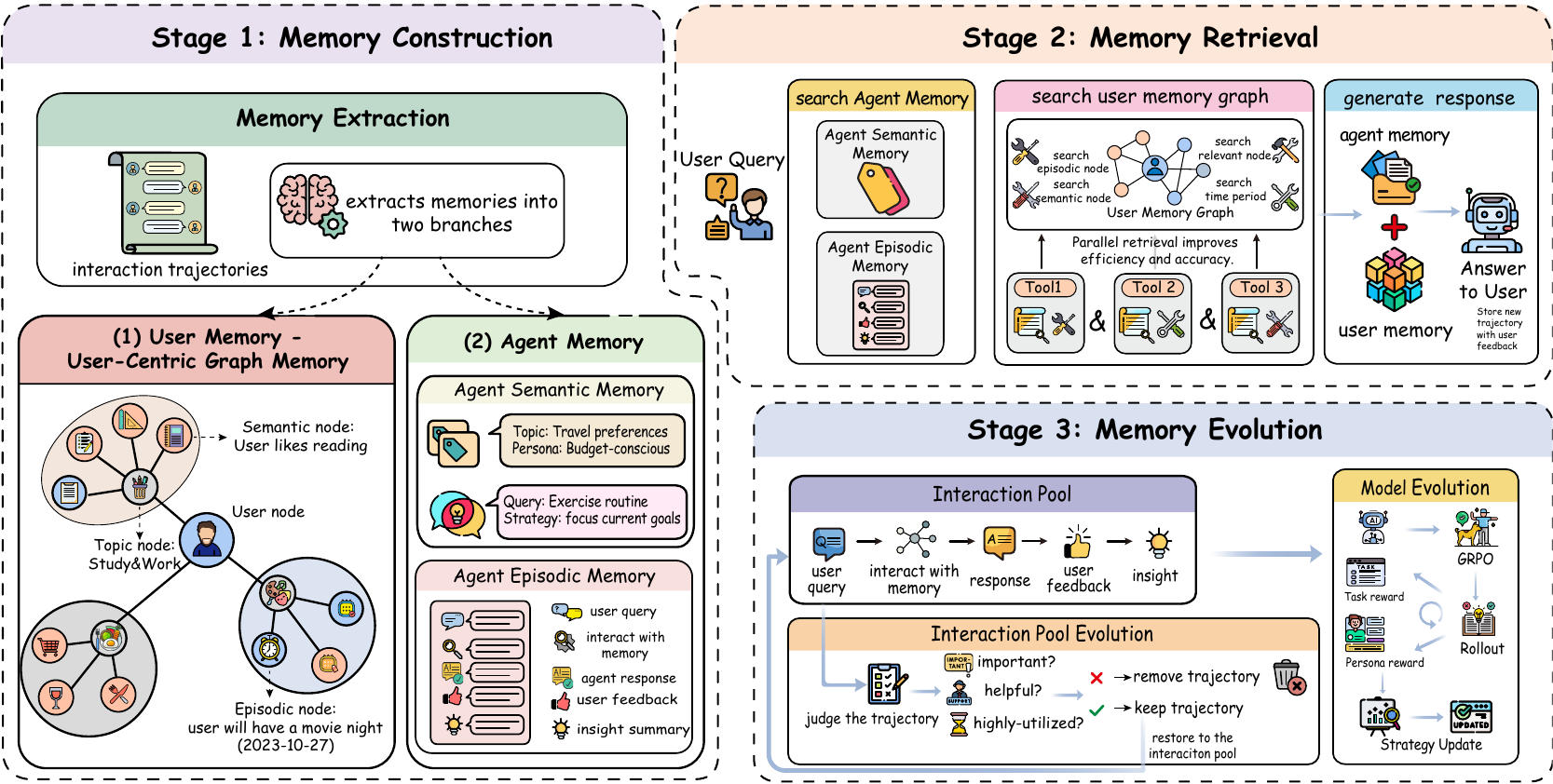}}
    \vspace{-4mm}
    \caption{Overview of PRISM. (1) Memory Construction extracts historical interactions into user memories and agent episodic memories. (2) Memory Retrieval combines retrieved agent memory with reasoning on the User Memory Graph for personalization. (3) Memory Evolution optimizes the interaction pool and updates the on-device LLM via reinforcement learning.}
    \label{fig:framework}
    \vspace{-4mm}
\end{figure*}

In this paper, we propose \textbf{PRISM}, a memory-based personalization framework with a dual-memory architecture centered on both the user and the agent, as illustrated in Figure~\ref{fig:framework}. PRISM consists of three interrelated stages: \textbf{Memory Construction}, \textbf{Memory Retrieval}, and \textbf{Memory Evolution}.
Given historical interactions, PRISM extracts \textit{user memories} and \textit{agent episodic memories}, storing them in the user memory graph and the interaction pool, respectively. During inference, the agent first retrieves relevant past experiences from agent memory, then performs retrieval over the user memory graph to generate personalized responses. In the memory evolution stage, PRISM filters low-value interaction units to maintain a high-quality and learnable interaction pool, which is further used to train the on-device LLM via reinforcement learning, enabling continuous adaptation to evolving user needs and task requirements.

\subsection{Memory Construction}
PRISM maintains a dual-memory architecture consisting of User Memory and Agent Memory, both of which contain episodic and semantic components but serve different roles.

User Memory captures user-related information through two components:
(1) User Episodic Memory ($\text{EM}_{\text{user}}$), which records concrete user experiences and contextual details; and
(2) User Semantic Memory ($\text{SM}_{\text{user}}$), which stores general user knowledge, preferences, and abstract concepts derived from historical data.

In contrast, Agent Memory guides the agent's retrieval and response strategies. It consists of:
(1) Agent Episodic Memory ($\text{EM}_{\text{agent}}$), which stores complete task-oriented interaction trajectories, including the user query, reasoning process, response, feedback, and experiential summary; and
(2) Agent Semantic Memory ($\text{SM}_{\text{agent}}$), which contains user personas and high-level strategies represented by insight signals extracted from agent episodic memories.

\paratitle{Interaction Decomposition.}
We logically categorize the interaction history $\mathcal{I}$ between the user and the agent into two distinct parts:
\begin{equation}
\mathcal{I} = \{ \mathcal{I}_{\text{query}}, \mathcal{I}_{\text{chat}} \},
\end{equation}
where $\mathcal{I}_{\text{query}}$ denotes task-oriented user queries, and
$\mathcal{I}_{\text{chat}}$ represents conversational interactions that convey user information and experiences.

\paratitle{User Memory Extraction.}
The agent $\mathcal{L}$ extracts user-related memories, including topic, semantic memory, and episodic memory, from $\mathcal{I}_{\text{chat}}$:
\begin{equation}
\langle t, s, e \rangle \leftarrow \mathcal{L}\!\left( \mathcal{I}_{\text{chat}} \right),
\end{equation}
where $t$ denotes the interaction topic, $s$ represents semantic memory, and $e$ corresponds to episodic memory. Implementation details are provided in Appendix~\ref{app:memory_construction}.

PRISM models user memory as a graph-based hierarchical structure $G$ rooted at a \texttt{UserNode}, where topic nodes organize semantic and episodic memory nodes.
Given the extracted tuple $(t,s,e)$, the agent maintains the graph through \textsc{Add}, \textsc{Update}, and \textsc{Delete} operations.
For example, if newly extracted information modifies a previously stored user preference, the agent updates the corresponding memory node rather than creating a separate duplicate:
\begin{equation}
\begin{gathered}
\text{Op} \in \{\textsc{Add}, \textsc{Update}, \textsc{Delete}\}, \\
G \leftarrow \text{Op}\big((t, s, e), G\big).
\end{gathered}
\end{equation}

\paratitle{Agent Episodic Memory Construction.}
We define a complete task-oriented interaction as:
\begin{equation}
\mathcal{I}_{\text{query}} = \langle Q, S, A, F \rangle,
\end{equation}
where $Q$ denotes the user query, $S$ represents the agent's internal reasoning and retrieval process, $A$ is the generated response, and $F$ represents user feedback.

Given $\mathcal{I}_{\text{query}}$, the agent further derives a topic $t$ and an insight signal $i$. 
Accordingly, agent episodic memory is defined as:
\begin{equation}
\text{EM}_{\text{agent}} = \langle Q, S, A, F, t, i \rangle.
\end{equation}

\paratitle{Agent Semantic Memory Construction.}
To construct agent semantic memory, PRISM extracts insight signals from agent episodic memories as high-level strategies and collects the corresponding user semantic memories as user personas.

Specifically, PRISM represents the high-level strategy under $t$ by retaining the insight signals from agent episodic memories associated with $t$:
\begin{equation}
\operatorname{Strategy}_{t}
=
\left\{
i_j
\mid
\text{EM}_{\text{agent}}^{(j)} \in \text{EM}_{\text{agent}},\,
t_j = t
\right\}.
\end{equation}
Here, $i_j$ denotes the insight field of $\text{EM}_{\text{agent}}^{(j)}$. 
Thus, the high-level strategy is directly represented by the insight signals retained from relevant agent episodic memories.

Meanwhile, user semantic memories associated with $t$ are collected as the corresponding user persona:
\begin{equation}
\text{SM}_{\text{user}}^{(t)} = \{ s_1^{(t)}, \ldots, s_n^{(t)} \},
\end{equation}

\begin{equation}
\operatorname{Persona}_{t}
=
\left\{
s_j
\mid
s_j \in \text{SM}_{\text{user}},\,
t_j = t
\right\}.
\end{equation}
Here, $t_j$ denotes the topic associated with semantic memory $s_j$. No additional aggregation operator is introduced.
Finally, agent semantic memory is defined as:
\begin{equation}
\text{SM}_{\text{agent}}
=
\langle
\operatorname{Strategy}_{t},
\operatorname{Persona}_{t}
\rangle .
\end{equation}

The resulting agent semantic memory combines high-level strategies and user personas to guide future retrieval and response generation.

\subsection{Memory Retrieval}
PRISM adopts a hybrid retrieval strategy that combines sequential and parallel mechanisms.
Specifically, the system first queries agent memory to obtain high-level guidance strategies and historical interaction experiences. 
These retrieved agent memories are then used to guide parallel exploration over the user memory graph. After retrieval, the language model synthesizes the collected memories and strategies to generate a personalized response. 
This retrieval mechanism preserves the complementary strengths of different memory components while enabling more comprehensive and personalized memory recall. 
We describe each step of the retrieval process in detail below.

\paratitle{Agent Memory Retrieval.}
Given a user query $u_i$ with topic $t_i$, PRISM first retrieves relevant memories from agent memory.
Formally, agent episodic memory retrieval is defined as:
\begin{equation}
\mathcal{R}^{\mathrm{AEM}}_i
=
\textsc{Retrieve}\!\left(u_i,{EM}_{\text{agent}}\right).
\end{equation}

Similarly, agent semantic memory is retrieved to provide strategy-level guidance:
\begin{equation}
\mathcal{R}^{\mathrm{ASM}}_i
=
\textsc{Retrieve}\!\left(u_i,{SM}_{\text{agent}}\right).
\end{equation}

\paratitle{User Memory Graph Retrieval.}
Conditioned on the retrieved agent memories $(\mathcal{R}^{\mathrm{AEM}}_i,\mathcal{R}^{\mathrm{ASM}}_i)$,
the LLM agent simultaneously generates multiple retrieval actions that can be executed in parallel over the user memory graph.
User memory retrieval is conducted over the memory graph $\mathcal{G}$ through a set of graph-based retrieval tools.

To ensure retrieval diversity and task adaptivity, PRISM exposes a set of memory retrieval tools to the LLM agent, including operations such as SearchRelated, SearchEntity, and other graph-based exploration functions.
These tools allow different retrieval branches to explore the user memory graph from complementary perspectives, such as related topics, entities, and neighboring memory nodes.

Conditioned on the retrieved agent memories 
$\left(\mathcal{R}^{\mathrm{AEM}}_i, \mathcal{R}^{\mathrm{ASM}}_i\right)$,
the agent performs multiple retrieval branches over the user memory graph.
Let $\mathcal{R}^{\mathrm{UM}}_{i,k}$ denote the user memories retrieved by the $k$-th retrieval branch.
The aggregated user memory context is then defined as:
\begin{equation}
\mathcal{R}^{\mathrm{UM}}_i
=
\bigcup_{k}
\mathcal{R}^{\mathrm{UM}}_{i,k}.
\end{equation}

Finally, the language model $\mathcal{L}$ synthesizes the gathered context to generate a personalized response.
By conditioning on the original query $u_i$, the retrieved agent memories $\mathcal{R}^{\mathrm{AEM}}_i$ and $\mathcal{R}^{\mathrm{ASM}}_i$, and the aggregated user memory context $\mathcal{R}^{\mathrm{UM}}_i$, the final response $a_i$ is generated as:
\begin{equation}
a_i = \mathcal{L}\left( u_i, \mathcal{R}^{\mathrm{AEM}}_i, \mathcal{R}^{\mathrm{ASM}}_i, \mathcal{R}^{\mathrm{UM}}_i \right).
\end{equation}

\subsection{Memory Evolution}
The memory evolution stage aims to continuously evolve both the interaction pool and the agent itself, enabling PRISM to adapt to dynamic task distributions and to improve personalization capabilities over time.
This stage consists of two processes: \textit{interaction pool evolution} and \textit{agent evolution}.

\paratitle{Interaction Pool Evolution.}
A naively growing interaction pool that only accumulates memories is insufficient for long-term deployment.
As task distributions evolve and model capabilities improve, unfiltered accumulation introduces redundancy and degrades retrieval efficiency.
To address this issue, PRISM performs dynamic interaction pool evolution through selective retention and removal.

For each agent episodic memory unit $\text{EM}_{\text{agent}}$, we define a multi-factor value vector:
\begin{equation}
V_{\text{EM}} = \langle T, I, L \rangle,
\end{equation}
where $T$ denotes temporal utility, $I$ represents importance, and $L$ measures learnability.
Temporal utility $T$ reflects both recency and retrieval frequency;
importance $I$ characterizes the intrinsic importance of a memory, indicating the extent to which it should be preserved for long-term decision-making and reasoning;
learnability $L$ measures the gain brought by reusing this memory, reflecting its positive contribution to performance improvement on subsequent tasks.

To enable threshold-based filtering, we convert $V_{\text{EM}}$ into a scalar quality score:
\begin{equation}
S_{\text{EM}} = \phi\!\left(V_{\text{EM}}\right),
\end{equation}
where $\phi(\cdot)$ is a deterministic aggregation function that maps $(T, I, L)$ to a scalar value.

We define a threshold-based removal rule, where a memory unit $\text{EM}_{\text{agent}}$ is discarded if its quality score satisfies $S_{\text{EM}} < \alpha$, with $\alpha$ denoting a predefined threshold. Implementation details are provided in Appendix~\ref{app:memory_evolution}.


\paratitle{Agent Evolution via Reinforcement Learning.}
While retrieval-time prompting and memory reuse can improve personalization, they are insufficient to induce intrinsic personalization behaviors within the model.
Correspondingly, in on-device large language model personalization scenarios, PRISM further optimizes the on-device LLM that serves as the agent's policy model via reinforcement learning.
The RL training process is grounded in the interaction pool, where we utilize user queries and corresponding feedback as direct supervision signals to optimize the model's policy.

\paratitle{Reward Design.}
We design a composite reward that jointly evaluates task correctness and personalization alignment. For a generated trajectory $\tau$, the reward is defined as:
\begin{equation}
R(\tau) = w_{\text{acc}}\, R_{\text{task}}(\tau)\; \text{+}\; w_{\text{persona}}\, R_{\text{user}}(\tau),
\end{equation}
where $w_{\text{acc}}$ and $w_{\text{persona}}$ are tunable coefficients that balance task accuracy and personalization alignment, respectively.
The task completion reward $R_{\text{task}}$ provides the primary learning signal by assessing whether the agent successfully completes the task, ensuring that the generated response is correct and usable at the content level.
The user persona reward $R_{\text{user}}$ is computed using an LLM-based evaluator $S_{\text{judge}}$, which evaluates the consistency between the generated response and the user feedback.
Specifically, $S_{\text{judge}}$ measures whether the response appropriately reflects user preferences, historical behaviors, and inferred stable traits.

\paragraph{Optimization Objective.}
In practice, we employ Group Relative Policy Optimization (GRPO)~\cite{shao2024deepseekmath} for agent optimization. Specifically, for each user query, we sample a group of $G$ candidate trajectories $\{\tau_i\}_{i=1}^{G}$ from the current policy.

The optimization objective is formulated as:
\begin{equation}
\mathcal{J}\text{(}\theta\text{)} = \mathbb{E} \left[ \frac{1}{G} \sum_{i=1}^{G} \rho_{\theta}^{\text{(}i\text{)}} A_i - \beta \mathbb{D}_{\mathrm{KL}}[\pi_\theta \| \pi_{\mathrm{ref}}] \right],
\end{equation}
where each candidate trajectory $i$ yields a reward $r_i$. The standardized group-relative advantage is computed as 
$A_i = \frac{r_i - \mathrm{mean}\text{(}\mathbf{r}\text{)}}{\mathrm{std}\text{(}\mathbf{r}\text{)}}$, 
with $\mathbf{r} = \{r_1, \dots, r_G\}$ representing the reward set, and $\rho_{\theta}^{\text{(}i\text{)}}$ denoting the per-action importance ratio. The KL divergence term regularizes the policy updates to prevent excessive deviation from the reference model $\pi_{\mathrm{ref}}$.

Through the iterative refinement of the interaction pool and RL-based agent optimization, PRISM facilitates a \textbf{dual-level adaptation process}. The interaction pool is continuously distilled to maintain a compact and high-quality experience repository, while the agent concurrently acquires adaptive, resource-aware, and memory-driven retrieval and response strategies. This synergy realizes the co-evolution of both the interaction pool and the agent itself.

\section{Experiments}
\subsection{Research Questions}

We evaluate PRISM from three perspectives: whether it improves long-term memory recall and utilization in long-horizon multi-turn interactions (RQ1), whether it constructs stable and evolving personalized memory representations from long-term user histories (RQ2), and whether it captures latent psychological traits for consistent personalized reasoning (RQ3).
\definecolor{bestcolor}{rgb}{0.741, 0.843, 0.933}  
\definecolor{secondcolor}{rgb}{0.886, 0.937, 0.855} 

\providecommand{\cc}{}
\renewcommand{\cc}[2]{%
  \colorbox{#1}{%
    \makebox[2.4em]{%
      \rule[-0.6ex]{0pt}{2.3ex}%
      #2%
    }%
  }%
}

\providecommand{\tc}{}
\renewcommand{\tc}[2]{%
  \colorbox{#1}{%
    \rule[-0.6ex]{0pt}{2.3ex}%
    #2%
  }%
}

\begin{table*}[t]
\centering
\caption{Performance comparison on Mem-PAL. The \tc{bestcolor}{best} and \tc{secondcolor}{second-best} results are highlighted.}
\begin{threeparttable}
\renewcommand{\arraystretch}{1.0}
\setlength{\tabcolsep}{2.5pt}

\resizebox{\linewidth}{!}{%
\begin{tabular}{ll ccc ccc cc ccc}
\toprule

\multirow{2}{*}{\textbf{Backbone}} & \multirow{2}{*}{\textbf{Method}} &
\multicolumn{3}{c}{\textbf{Requirement Restatement}} &
\multicolumn{3}{c}{\textbf{Solution Proposal}} &
\multicolumn{2}{c}{\textbf{Multi-turn dialogue}} &
\multicolumn{3}{c}{\textbf{Average}} \\

\cmidrule(lr){3-5} \cmidrule(lr){6-8} \cmidrule(lr){9-10} \cmidrule(lr){11-13}
 &  & B1 & ACC & J1 & B1 & ACC & J1 & B1 & J1 & B1 & ACC & J1 \\
\midrule

\multirow{5}{*}{GPT-5-mini}
 & Base           & 13.53 & 17.10 & 21.88 & 14.04 & 27.54 & 28.96 & 19.71 & 27.84 & 14.76 & 22.32 & 26.23 \\
 & LangMem        & 16.41 & 24.94 & 28.06 & 14.12 & 39.47 & 41.83 & 19.20 & 34.05 & 15.58 & 32.20 & 34.65 \\
 & Mem0 (entity)  & \cc{secondcolor}{22.96} & \cc{secondcolor}{26.42} & \cc{secondcolor}{30.96} & 18.83 & \cc{secondcolor}{46.12} & \cc{secondcolor}{50.36} & \cc{secondcolor}{25.15} & \cc{secondcolor}{39.43} & 20.68 & \cc{secondcolor}{36.27} & \cc{secondcolor}{40.25} \\
 & A-MEM          & 21.34 & 25.72 & 29.18 & \cc{secondcolor}{19.67} & 41.41 & 46.88 & 23.08 & 37.26 & \cc{secondcolor}{21.03} & 33.56 & 37.77 \\
 & PRISM          & \cc{bestcolor}{23.65} & \cc{bestcolor}{37.71} & \cc{bestcolor}{44.31} & \cc{bestcolor}{20.72} & \cc{bestcolor}{51.39} & \cc{bestcolor}{56.93} & \cc{bestcolor}{26.18} & \cc{bestcolor}{46.52} & \cc{bestcolor}{23.18} & \cc{bestcolor}{44.55} & \cc{bestcolor}{49.25} \\
\midrule

\multirow{7}{*}{Qwen2.5-7B-Instruct}
 & Base+RAG       & 12.21 & 10.87 & 15.29 & 13.62 & 13.01 & 17.22 & 20.82 & 18.01 & 15.55 & 11.94 & 16.84 \\
 & LangMem        & 14.02 & 19.58 & 24.12 & 17.58 & 20.10 & 23.97 & 19.21 & 20.31 & 16.94 & 19.84 & 22.80 \\
 & Mem0 (entity)  & 19.15 & 22.85 & 25.34 & 20.49 & 21.01 & 26.03 & 21.04 & 26.15 & 20.30 & 21.93 & 25.84 \\
 & A-MEM          & 17.83 & 23.67 & 27.03 & 19.45 & 21.31 & 25.86 & 20.52 & 23.61 & 19.27 & 22.49 & 25.50 \\
 & Mem-$\alpha$(RL)  & 20.18 & 27.06 & 30.51 & 20.53 & 24.15 & 30.17 & 21.64 & 27.93 & 20.78 & 25.60 & 29.54 \\
 & PRISM          & \cc{secondcolor}{22.52} & \cc{secondcolor}{30.39} & \cc{secondcolor}{33.99} & \cc{secondcolor}{21.61} & \cc{secondcolor}{26.94} & \cc{secondcolor}{34.50} & \cc{secondcolor}{22.15} & \cc{secondcolor}{32.27} & \cc{secondcolor}{22.29} & \cc{secondcolor}{28.66} & \cc{secondcolor}{33.59} \\
 & PRISMRL        & \cc{bestcolor}{23.39} & \cc{bestcolor}{34.69} & \cc{bestcolor}{38.50} & \cc{bestcolor}{22.05} & \cc{bestcolor}{30.09} & \cc{bestcolor}{37.38} & \cc{bestcolor}{22.75} & \cc{bestcolor}{37.08} & \cc{bestcolor}{22.53} & \cc{bestcolor}{32.39} & \cc{bestcolor}{37.65} \\

\bottomrule
\end{tabular}%
}
\begin{tablenotes}[flushleft]
        \footnotesize 
        \item Note: B1, ACC, and J1 denote BLEU-1, Accuracy, and the LLM-as-a-Judge score, respectively. 
        The average scores are computed\\ separately within each metric across the corresponding task types; different metrics are not averaged together.
\end{tablenotes}
\end{threeparttable}

\label{tab:mempal}
\end{table*}

\definecolor{bestcolor}{rgb}{0.741, 0.843, 0.933}   
\definecolor{secondcolor}{rgb}{0.886, 0.937, 0.855} 

\providecommand{\cc}{}
\renewcommand{\cc}[2]{%
  \colorbox{#1}{%
    \makebox[2.4em]{%
      \rule[-0.7ex]{0pt}{2.5ex}%
      #2%
    }%
  }%
}

\providecommand{\tc}{}
\renewcommand{\tc}[2]{%
  \colorbox{#1}{%
    \rule[-0.7ex]{0pt}{2.5ex}%
    #2%
  }%
}

\begin{table*}[t]
\centering
\caption{Performance comparison on KnowMeBench. The \tc{bestcolor}{best} and \tc{secondcolor}{second-best} results are highlighted.}
\begin{threeparttable}
\renewcommand{\arraystretch}{1.05}
\setlength{\tabcolsep}{2.5pt}

\resizebox{\linewidth}{!}{%
\begin{tabular}{ll cccc ccc ccc c}
\toprule
\multirow{2}{*}{\textbf{Backbone}} & \multirow{2}{*}{\textbf{Method}} &
\multicolumn{4}{c}{\textbf{Level I: Fact \& Entity}} &
\multicolumn{3}{c}{\textbf{Level II: Temporal Logic}} &
\multicolumn{3}{c}{\textbf{Level III: Insight}} &
\multirow{2}{*}{\textbf{Overall}} \\

\cmidrule(lr){3-6} \cmidrule(lr){7-9} \cmidrule(lr){10-12}

 &  & T1 & T2 & T3 & \textbf{Avg.} & T4 & T5 & \textbf{Avg.} & T6 & T7 & \textbf{Avg.} & \textbf{Avg.} \\
\midrule

\multirow{5}{*}{GPT-5-mini} 
 & Base          & 76.97 & 71.93 & 53.58 & 67.50 & 55.84 & 43.82 & 49.83 & 36.04 & 30.11 & 33.07 & 51.54 \\
 & LangMem       & 80.91 & \cc{secondcolor}{77.27} & 55.22 & 71.13 & 57.44 & 42.24 & 49.84 & 37.49 & 29.68 & 33.59 & 52.99 \\
 & Mem0 (entity) & 81.88 & 71.40 & \cc{secondcolor}{56.22} & 69.83 & 57.92 & \cc{secondcolor}{44.41} & 51.16 & \cc{secondcolor}{39.11} & \cc{secondcolor}{32.87} & \cc{secondcolor}{35.99} & 53.71 \\
 & A-MEM         & \cc{secondcolor}{82.61} & 76.60 & 56.08 & \cc{secondcolor}{71.76} & \cc{secondcolor}{59.04} & 43.68 & \cc{secondcolor}{51.36} & 38.71 & 32.34 & 35.52 & \cc{secondcolor}{54.15} \\
 & PRISM         & \cc{bestcolor}{85.03} & \cc{bestcolor}{85.93} & \cc{bestcolor}{61.00} & \cc{bestcolor}{77.32} & \cc{bestcolor}{68.32} & \cc{bestcolor}{55.70} & \cc{bestcolor}{62.01} & \cc{bestcolor}{48.36} & \cc{bestcolor}{44.57} & \cc{bestcolor}{46.47} & \cc{bestcolor}{62.51} \\
\midrule

\multirow{7}{*}{Qwen2.5-7B-Instruct}
 & Base+RAG       & 56.30 & 50.87 & 33.62 & 46.93 & 35.36 & 30.71 & 33.03 & 22.18 & 18.51 & 20.35 & 34.66 \\
 & A-MEM          & 57.15 & 54.40 & 38.32 & 49.96 & 35.04 & 31.46 & 33.25 & 23.64 & 18.72 & 21.18 & 36.65 \\
 & Mem0 (entity)  & 56.91 & 55.27 & 40.11 & 50.76 & 35.68 & 31.74 & 33.71 & 23.92 & 19.15 & 21.53 & 37.29 \\
 & LangMem        & 56.97 & 58.80 & 38.25 & 51.34 & 38.40 & 32.19 & 35.29 & 24.81 & 20.53 & 22.67 & 37.81 \\
 & Mem-$\alpha$(RL) & 58.61 & 60.13 & \cc{secondcolor}{45.49} & 54.74 & 40.96 & 30.50 & 35.73 & 29.41 & 29.15 & 29.28 & 41.00 \\
 & PRISM          & \cc{secondcolor}{60.24} & \cc{secondcolor}{62.13} & 45.06 & \cc{secondcolor}{55.81} & \cc{secondcolor}{49.12} & \cc{secondcolor}{34.80} & \cc{secondcolor}{41.96} & \cc{secondcolor}{32.04} & \cc{secondcolor}{31.38} & \cc{secondcolor}{31.71} & \cc{secondcolor}{43.38} \\
 & PRISMRL        & \cc{bestcolor}{67.82} & \cc{bestcolor}{73.07} & \cc{bestcolor}{55.04} & \cc{bestcolor}{65.31} & \cc{bestcolor}{52.96} & \cc{bestcolor}{45.27} & \cc{bestcolor}{49.11} & \cc{bestcolor}{33.33} & \cc{bestcolor}{35.85} & \cc{bestcolor}{34.59} & \cc{bestcolor}{50.91} \\
\bottomrule
\end{tabular}%
}

\begin{tablenotes}[flushleft]
        \footnotesize 
        \item Note: T1 to T7 denote Context-Aware Extraction, Adversarial Abstention, Temporal Reasoning, Logical Event Ordering, Mnestic \\ Trigger Analysis, Mind-Body Interaction, and Expert-Annotated Psychoanalysis, respectively.
\end{tablenotes}
\end{threeparttable}
\label{tab:knowme}
\end{table*}

\subsection{Experimental Setup}
\paratitle{Evaluation Datasets.}
To empirically address the research questions above, the evaluation datasets should cover both long-term memory utilization and personalized reasoning. Since traditional memory benchmarks solely focus on information recall accuracy \cite{maharana2024evaluating}, we adopt Mem-PAL \cite{huang2025mem} and KnowMeBench \cite{wu2026knowme} as our main evaluation datasets. Mem-PAL focuses on personalized memory construction and utilization across long-term, multi-session interactions, while KnowMeBench further evaluates latent user preferences and psychological traits reflected in personalized reasoning. We also conduct supplementary experiments on LoCoMo \cite{maharana2024evaluating}, with details provided in Appendix~\ref{app:supplementary}.

\paratitle{Evaluation Metrics.}
For Mem-PAL, we utilize BLEU-1, Accuracy (Acc), and $J_1$ to multi-dimensionally evaluate the overall quality and user preferences of the memory tasks. For KnowMeBench, we primarily rely on $J_1$ to conduct a comprehensive evaluation of response accuracy and personalization capabilities. For LLM-as-a-Judge evaluation, we use GPT-4o as the judge model.

\paratitle{Baselines and LLM Backbones.}
We compare PRISM with four representative memory-augmented agent systems, namely Mem0 \cite{mem0}, LangMem \cite{langmem2024github}, A-Mem \cite{xu2025mem}, and Mem-Alpha \cite{wang2025mem}. The backbone language models used in our experiments include GPT-5-mini and Qwen2.5-7B-Instruct. For Qwen2.5-7B-Instruct, a base model combined with retrieval-augmented generation (Base Model + RAG) is adopted to ensure basic task-solving capability.

\paratitle{Implementation Details.}
We use Neo4j to store structured user memories and Qdrant to store agent memories, and fine-tune PRISM using the \texttt{verl} framework~\cite{sheng2025hybridflow}. Additional experimental details are provided in Appendix~\ref{app:dataset}.

\begin{figure*}[t]
    \centering
    \includegraphics[width=\textwidth]{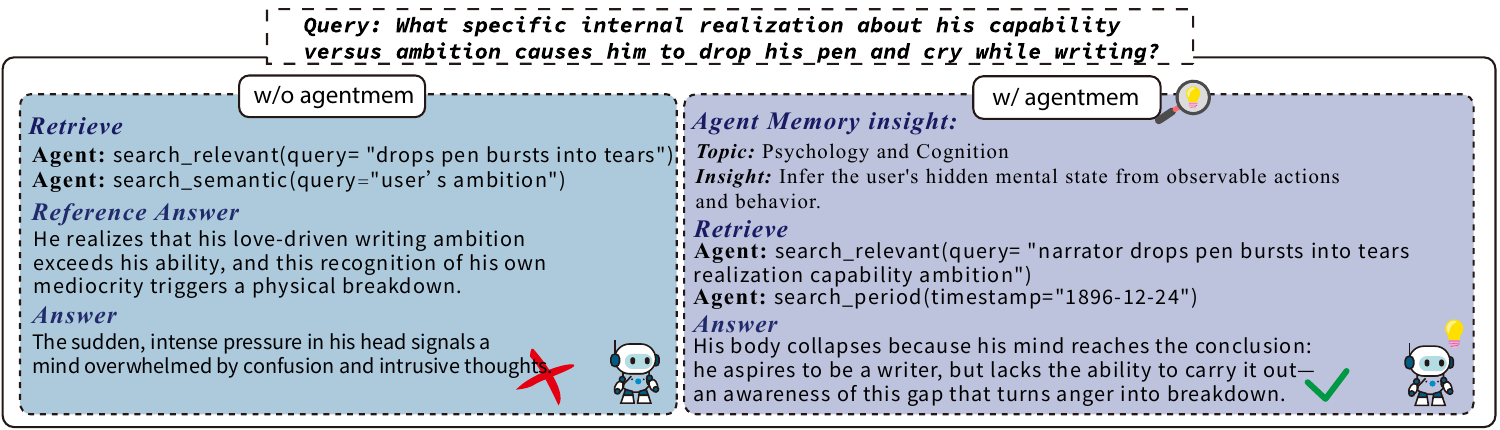}
    \caption{Qualitative comparison with and without agent memory, illustrating how agent memory helps guide retrieval toward relevant evidence and supports more context-aware reasoning in a psychological inference case.}
\label{fig:agentmemory}
    \vspace{-5mm}
\end{figure*}
\subsection{Main Result Analysis}
\paratitle{Effectiveness in Long-term Memory Retrieval.}
To evaluate memory retrieval and utilization (RQ1), we analyze KnowMeBench Level I (Fact \& Entity) and Level II (Temporal Logic) results in Table~\ref{tab:knowme}. PRISM shows a stable advantage in Level I and Level II tasks. On GPT-5-mini, PRISM improves over Mem0 by 7.49 points on Level I and 10.85 points on Level II. On Qwen2.5-7B-Instruct, PRISM improves over Mem0 by 5.05 points on Level I and 8.25 points on Level II, while PRISMRL further enlarges the gains to 14.55 and 15.40 points, respectively. This suggests that, compared with Mem0's flat and entity-centric memory mechanism, PRISM better captures temporal and logical relationships for long-term memory reasoning.

\paratitle{Personalized Memory Modeling.}
To evaluate personalized memory modeling (RQ2), we analyze results on Mem-PAL, as shown in Table~\ref{tab:mempal}. Mem-PAL tests whether a model can infer users' implicit needs from long-term interaction histories and maintain coherent user profiles across multi-turn dialogues. For GPT-5-mini, PRISM improves the average J1 score from 40.25 to 49.25. For Qwen2.5-7B-Instruct, PRISMRL improves the average J1 score from 29.54 to 37.65. The gains are particularly clear in multi-turn settings, where PRISM and PRISMRL improve J1 by 7.09 and 9.15 points, respectively. This suggests that PRISM not only recalls static user attributes, but also maintains an evolving user persona for stable personalized reasoning.

\paratitle{Modeling Cognitive Patterns and Latent Traits.}
For RQ3, we evaluate PRISM on KnowMeBench Level III (Insight), which requires inferring latent psychological traits beyond explicit fact retrieval. As shown in Table~\ref{tab:knowme}, PRISM improves over the strongest non-PRISM baseline by 10.48 points under GPT-5-mini, while PRISMRL improves the Level III average from 29.28 to 34.59 under Qwen2.5-7B-Instruct. This suggests that strategy-guided reasoning and multi-dimensional memory retrieval help PRISM connect observable behaviors with deeper cognitive patterns, supporting more consistent cognitive personalization.

\paratitle{Summary.}
Overall, PRISM brings consistent improvements across factual recall, temporal reasoning, and cognitive insight tasks. The gains are especially clear in temporal logic and open-ended insight settings, showing that PRISM improves not only memory retrieval accuracy, but also the structural utilization of memory for personalized reasoning. We further evaluate PRISM on smaller backbones to examine its applicability to on-device settings, with detailed results provided in Appendix~\ref{app:supplementary}.

\subsection{Ablation Study}
\begin{figure}[t]
    \centering
    \resizebox{1\columnwidth}{!}{
    \includegraphics{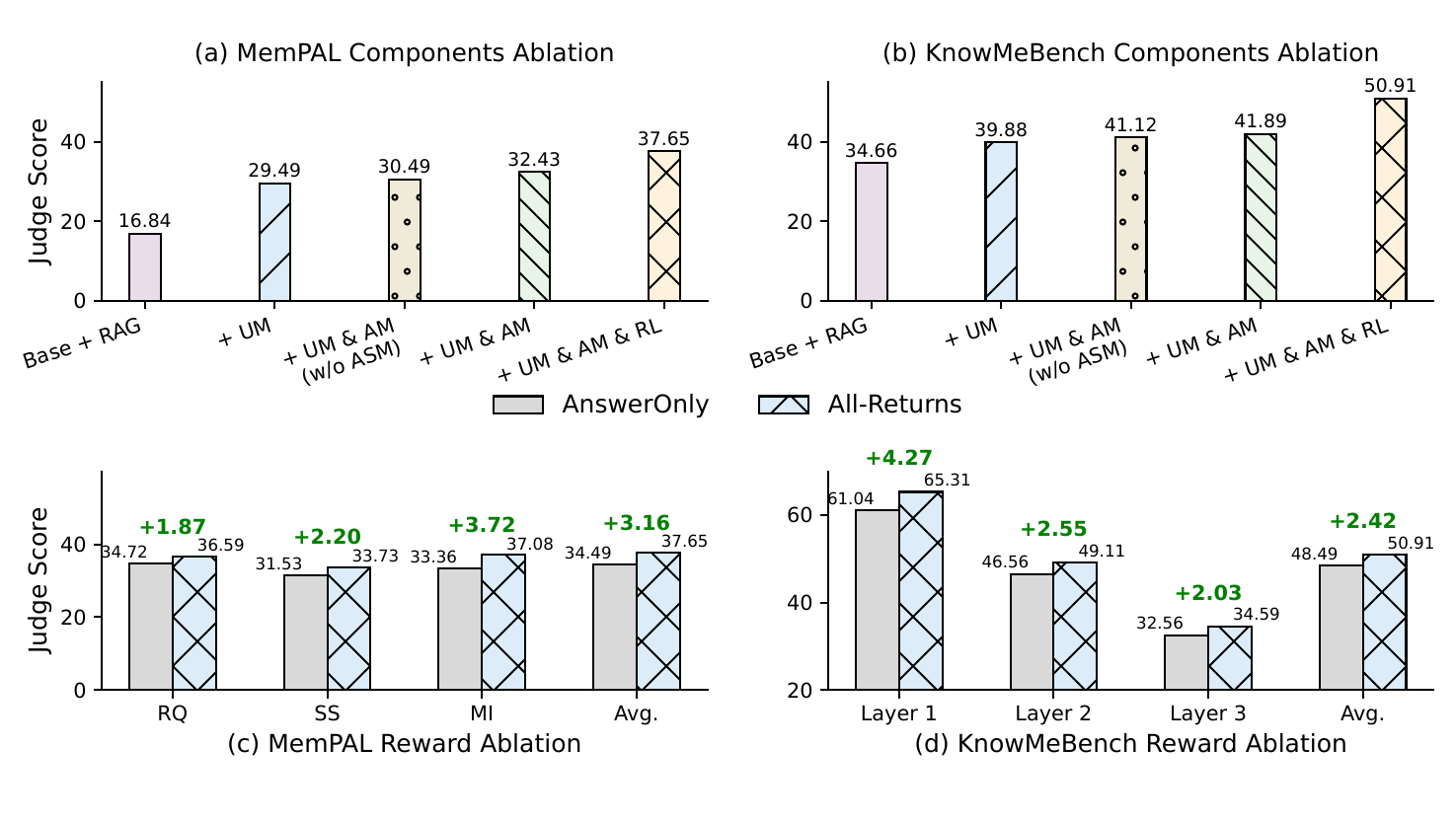}}
    \caption{Ablation analysis of PRISM's memory components and reward designs.}
    \label{fig:abl}
    \vspace{-7mm}
\end{figure}

\subsubsection{Memory Component Analysis}
To evaluate the contribution of PRISM components, we conduct ablation studies on user memory, agent memory, and reinforcement learning (RL) training. Figure~\ref{fig:abl} reports the results on Mem-PAL and KnowMeBench using Qwen2.5-7B-Instruct as the backbone model.

\paratitle{Effect of User Memory.}
Introducing the User Memory Graph substantially improves performance over Base+RAG, raising the score from 16.84 to 29.49 on Mem-PAL and from 34.66 to 39.88 on KnowMeBench. This shows that PRISM benefits from tool-based graph retrieval, which accesses temporal, semantic, and entity-related memory facets beyond simple vector matching.

\paratitle{Effect of Agent Memory.}
Agent memory guides the model in selecting retrieval paths and response strategies based on accumulated interaction experiences. As shown in Figure~\ref{fig:agentmemory}, it retrieves a domain-specific high-level strategy that helps the model formulate more precise search queries and locate critical evidence. This enables the model to reason beyond superficial observations and identify deeper causes behind user-related events or behaviors. The comparison between ``+ UM \& AM (w/o ASM)'' and ``+ UM \& AM'' further shows the contribution of agent semantic memory.

\paratitle{Effect of Reinforcement Learning.}
To evaluate the contribution of reinforcement learning, we compare the full PRISM model with a variant trained without RL. As shown in Figure~\ref{fig:abl}(a) and Figure~\ref{fig:abl}(b), adding RL improves the score from 32.43 to 37.65 on Mem-PAL and from 41.89 to 50.91 on KnowMeBench. This suggests that feedback-driven optimization helps the agent move beyond static memory storage and more effectively leverage user and agent memory for personalized response generation.
\subsubsection{Reward Function Analysis}
To evaluate the reward design, we compare the full multi-component reward strategy (All-Returns) with a variant using only task accuracy rewards (AnswerOnly). As shown in Figure~\ref{fig:abl}(c) and Figure~\ref{fig:abl}(d), All-Returns consistently outperforms AnswerOnly, improving the average score by 3.16 points on Mem-PAL and 2.42 points on KnowMeBench. This indicates that incorporating user persona rewards provides additional personalization benefits beyond task-accuracy-only optimization. We further study the weighting between user persona and task accuracy rewards, with detailed results provided in Appendix~\ref{app:supplementary}.

\section{Conclusion}

In this paper, we propose PRISM, a Personalized Dual-Memory Self-Evolving Framework for LLM-based agents. 
By jointly modeling user memory and agent memory, PRISM captures long-term interaction patterns and leverages agent-level experiences to guide memory retrieval and response generation. 
Furthermore, PRISM incorporates user feedback into reinforcement learning to improve both reasoning and personalization. 
Experiments on public benchmarks demonstrate that PRISM consistently outperforms existing memory-based baselines in response quality and personalized alignment. 

\section*{Limitations}

Although PRISM achieves strong performance on personalized memory benchmarks, several limitations remain. First, its self-evolution process relies on user feedback to refine memory retrieval and response strategies. However, in real-world scenarios, users may not always provide explicit, frequent, or reliable feedback after each interaction, which may limit the stability of feedback-driven optimization. Future work can explore more robust implicit feedback modeling and more adaptive optimization mechanisms.

\section*{Ethical Considerations}

PRISM stores and uses user-related information across long-term interactions, which raises concerns about data controllability, interpretability, and safety. To reduce privacy risks, PRISM organizes user memories as a structured memory graph, allowing user-related information to be located, inspected, updated, and deleted. This design helps preserve users' control over personal data and reduces the influence of outdated or inaccurate memories. Personalized memory systems may also face the risk of over-personalization, where agents rely too heavily on historical preferences or inferred traits. PRISM mitigates this risk through dynamic memory updating and memory evolution, while future deployments should further strengthen transparent memory management and safety safeguards.

\section*{Acknowledgments}
This work was supported by the Nanjing University-China Mobile Joint Research Institute.

\clearpage
\bibliography{sample-base}

\clearpage
\appendix
\section{Implementation Details of PRISM}
\label{app:implementation}

This appendix provides the implementation details of the PRISM framework.
We first illustrate the personalized memory storage and retrieval process through a concrete example (Appendix~\ref{app:PRISMdesign}). 
We then introduce the functions, variables, and a subset of core tools involved in this process, and present the core algorithmic formulation of the framework (Appendix~\ref{app:algorithms}). 
Finally, we provide a comparative case study with Mem0 to demonstrate the personalization capability and response accuracy of the PRISM framework (Appendix~\ref{app:case_study}).

\subsection{System Execution Workflow}
\label{app:workflow}

\begin{figure*}[t]
    \centering
    \includegraphics[width=\textwidth]{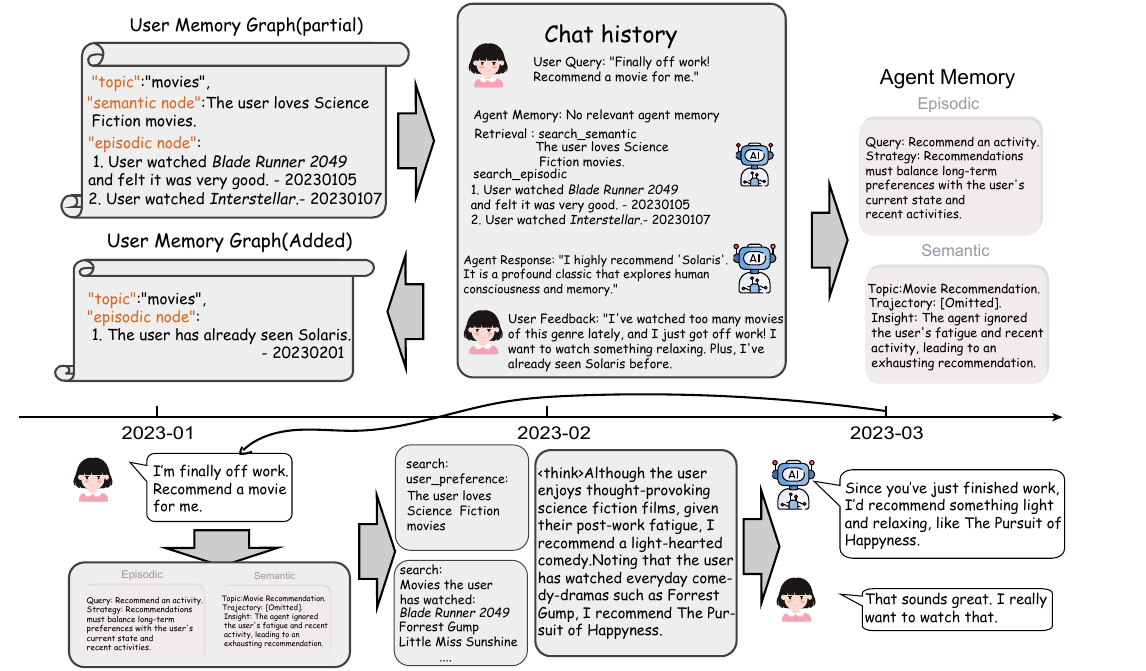}
    \caption{
    An example illustrating how PRISM learns from user feedback.
    The agent converts an unsatisfactory recommendation into episodic and semantic agent memory, and later uses this strategy-level experience together with the user memory graph to produce a more context-aware personalized recommendation.
    }
    \label{fig:Workflow}
    \vspace{-2mm}
\end{figure*}

As illustrated in Fig.~\ref{fig:Workflow}, we present a concrete execution trace of personalized movie recommendation. This example illustrates how PRISM performs personalized recommendation through a multi-step reasoning process that interleaves memory construction, retrieval, and action execution, while leveraging interaction feedback to support continual system evolution.

\paragraph{(1) Memory construction and abstraction (Alg.~\ref{alg:user_memory} and Alg.~\ref{alg:agent_memory}).}
Given the raw interaction text $T$, Alg.~\ref{alg:user_memory} updates the user memory graph $G_{user}$ by constructing topic nodes (e.g., \texttt{movies}), extracting semantic memories that represent long-term preferences (such as a preference for science-fiction content), and recording episodic memories that capture concrete interaction histories, including previously consumed items.
When the agent produces an initially misaligned recommendation—such as suggesting cognitively demanding or already-consumed content despite the user being in a post-work fatigued state—Alg.~\ref{alg:agent_memory} stores this interaction as an episodic trajectory in the agent memory store.
The agent then abstracts this experience into agent semantic memory, inducing a reusable high-level strategy indicating that recommendation decisions should balance long-term preferences with the user’s current state and recent activities.

\paragraph{(2) Retrieval-augmented parallel reasoning and response (Alg.~\ref{alg:retrieval}).}
In subsequent interactions, the system invokes the \textsc{ReasonAndRetrieve} procedure defined in Alg.~\ref{alg:retrieval}.
This procedure combines strategy signals from agent memory with personalized evidence from the user memory graph through retrieval-augmented reasoning to support the final decision.

During inference, the reasoning module first performs an internal planning phase to determine what information is required for the current decision, such as whether long-term preferences, recent activities, or immediate user state signals should be prioritized.
Conditioned on this plan, the agent retrieves persona and strategy signals from agent semantic memory $S_{agent}$ and simultaneously generates multiple retrieval actions to query the user memory graph $G_{user}$ in parallel, obtaining user-specific evidence such as watched history and contextual cues.

Based on the information obtained through parallel retrieval, the agent synthesizes memory evidence from different sources and generates an appropriate response, for example by adjusting the recommendation direction toward lighter and more relaxing content.
The generated response, together with the user's feedback, is appended to the interaction history and subsequently serves as the basis for memory updates and system evolution.

\paragraph{(3) Interaction pool and model evolution (Alg.~\ref{alg:pool_evolution} and Alg.~\ref{alg:model_evolution}).}
According to Alg.~\ref{alg:pool_evolution}, each episodic memory $\mathrm{EM}_j$ is evaluated using a composite score that accounts for temporal utility, importance, and learnability.
Interaction trajectories that contain explicit user corrections are assigned higher learnability scores and are therefore retained in the interaction pool $\mathcal{P}$ as valuable learning signals rather than being discarded.
Following Alg.~\ref{alg:model_evolution}, the learning process samples episodic memories from $\mathcal{P}$ and generates groups of candidate trajectories under the current policy $\pi_\theta$.
Rewards are computed from user feedback and extracted insight signals, and group-relative advantages are applied to update the policy via GRPO.
Through this optimization process, the policy is progressively biased toward behaviors that better align with user feedback, ensuring that for future queries under similar contextual conditions, the agent prioritizes the user’s immediate state instead of rigidly adhering to static long-term preferences.

\subsection{Detailed Design and Implementation}
\label{app:PRISMdesign}

This subsection presents the detailed design and implementation of the PRISM framework, focusing on the practical realization of the memory modules.

\subsubsection{Memory Construction}
\label{app:memory_construction}

In this subsection, we describe how different types of interaction content are stored in distinct memory modules, thereby providing structured and specialized representations for both user memory and agent memory.

\paratitle{User Memory Graph Node Definition.}
Following the definition in the main text, user-related memories are extracted as
\begin{equation}
\langle t, s, e \rangle \leftarrow \mathcal{L}\!\left( \mathcal{I}_{\text{chat}} \right).
\end{equation}

We next describe the definitions of the corresponding memory nodes in the user memory graph.
Topic nodes serve as the organizational backbone of the user memory graph and are hierarchically structured to support different levels of abstraction.
Each topic node represents an interaction theme inferred from user conversations and may have a parent topic, forming a tree-like hierarchy.
High-level topics capture coarse-grained attributes, while lower-level topics represent more fine-grained subtopics.
Formally, a topic node is defined as
\begin{equation}
t = \langle c_t, p_t \rangle,
\end{equation}
where $c_t$ denotes the semantic content of the topic, and $p_t$ represents its parent topic node.
If $p_t = \varnothing$, the topic is treated as a root-level topic.

\vspace{0.5em}

Semantic memory nodes encode stable and abstract attributes associated with a given topic, such as preferences, habits, or long-term traits.
Each semantic memory node is explicitly anchored to a topic node and is represented as a structured triple.
The head entity of a semantic memory is not restricted to the user and may correspond to any relevant subject entity inferred from the interaction.
Formally, a semantic memory node is defined as
\begin{equation}
s = \langle h, r, o, t \rangle,
\end{equation}
where $h$ denotes the head entity, $r$ represents a relation type, $o$ corresponds to the semantic attribute or object, and $t$ indicates the associated topic node.
Semantic memory nodes do not directly connect to episodic memory nodes and are accessed through their shared topic context.

\vspace{0.5em}

Episodic memory nodes record concrete interaction events and user experiences under a specific topic, preserving temporal and causal information.
Each episodic memory node is associated with exactly one topic node and may exhibit hierarchical or causal relations with other episodic nodes under the same topic.
Specifically, episodic memories can form parent--child structures that reflect event decomposition, as well as causal links that capture dependencies between events.
The segmentation and structuring of episodic memories are determined by an LLM-based extractor.
Formally, an episodic memory node is defined as
\begin{equation}
e = \langle d, p_e, c_e, t \rangle,
\end{equation}
where $d$ represents the event description, $p_e$ denotes an optional parent episodic node, $c_e$ indicates causal relations to other episodic events, and $t$ specifies the associated topic node.

\paratitle{Memory Graph Operations.}
For semantic and episodic memory nodes, when a new memory candidate is extracted, the system first retrieves the top-$k$ most relevant existing memories under the same topic based on semantic similarity.
Formally, given a new memory candidate $x \in \{s, e\}$ associated with topic $t$, similarity-based retrieval is performed as
\begin{equation}
\mathcal{R}_k(x) = \operatorname{TopK}\big(\operatorname{Sim}(x, \mathcal{M}_t)\big),
\end{equation}
where $\mathcal{M}_t$ denotes the set of existing semantic or episodic memory nodes under topic $t$, and $\operatorname{Sim}(\cdot)$ is a similarity function implemented via embedding similarity.

The retrieved memories $\mathcal{R}_k(x)$, together with the new memory candidate $x$, are then provided to an LLM-based judge to determine whether $x$ corresponds to an existing memory node.
If the new candidate is judged to represent previously unseen information, a new node is added to the graph.
If it is determined to refer to the same underlying memory while providing additional or refined information, the corresponding existing node is updated.
If the new candidate is deemed redundant or contradictory with respect to existing memories, it is discarded and no modification to the graph is performed.

For episodic memory nodes, the system further evaluates potential causal relations between the new event and existing episodic memories under the same topic.
Specifically, the LLM-based judge assesses whether the new episodic memory is causally dependent on or triggered by a prior event.
When such a causal relation is identified, the new episodic node is attached to the corresponding parent event via a directed causal edge, thereby encoding temporal and causal dependencies between events.

\vspace{0.5em}

Topic nodes are managed through a different procedure.
Given a newly inferred topic candidate $t'$, the system first retrieves the most similar existing topic nodes based on semantic similarity:
\begin{equation}
\mathcal{R}_k(t') = \operatorname{TopK}\big(\operatorname{Sim}(t', \mathcal{T})\big),
\end{equation}
where $\mathcal{T}$ denotes the set of existing topic nodes.

If no sufficiently similar topic node is identified, the candidate topic $t'$ together with the retrieved topics $\mathcal{R}_k(t')$ are jointly provided to the LLM-based judge.
Based on this comparison, the judge determines whether the new topic should replace an existing topic node, be retained as an independent topic, or be attached as a subtopic under an existing node.
This mechanism allows the topic hierarchy to evolve dynamically while preserving structural consistency within the user memory graph.

\paratitle{Agent Episodic Memory Storage.}

After an agent episodic memory is extracted as
\begin{equation}
\text{EM}_{\text{agent}} = \langle Q, S, A, F, t, i \rangle,
\end{equation}
the system performs similarity-based retrieval over the agent episodic memory store to identify past interactions with similar user queries.
Formally, given the current query $Q$, we retrieve the top-$k$ episodic memories as
\begin{equation}
\mathcal{R}_k(Q) = \operatorname{TopK}\big(\operatorname{Sim}(Q, \mathcal{Q}_{\text{agent}})\big),
\end{equation}
where $\mathcal{Q}_{\text{agent}}$ denotes the set of stored queries in agent episodic memory and $\operatorname{Sim}(\cdot)$ is an embedding-based similarity function.

Given the retrieved set $\mathcal{R}_k(Q)$, the system determines whether the current query corresponds to previously observed query patterns.
If similar queries are identified, their shared semantic characteristics are abstracted, and representative user tendencies are extracted from the associated interaction records and feedback signals.
Based on this abstraction, an insight signal is summarized as
\begin{equation}
i = \textsc{Summarize}\big(\{Q_j, S_j, A_j, F_j\}_{Q_j \in \mathcal{R}_k(Q)}\big),
\end{equation}
where $i$ captures the user’s implicit preferences, response expectations, and feedback-supported experience under topic $t$.

The insight signal provides a compact abstraction of the retrieved episodic memories and is used to guide future strategy retrieval and agent behavior.
This design reduces redundant reasoning over similar past interactions while preserving user-aligned signals essential for future personalization.

\paratitle{Agent Semantic Memory Storage.}

This section details the operational storage mechanisms that bridge the theoretical definitions of agent episodic and semantic memories.
The storage architecture functions as a hierarchical structure: agent episodic memory serves as a comprehensive archival system, whereas agent semantic memory serves as a distilled knowledge base for high-level guidance.

Specifically, Agent Episodic Memory prioritizes \textit{contextual fidelity} by retaining the complete interaction tuple $\langle Q, S, A, F, t, i \rangle$.
While the full tuple ensures the traceability of the reasoning process ($S$) and generated response ($A$), the primary signal for system evolution is the insight ($i$).
This insight, derived from recurring patterns and feedback, acts as a condensed information carrier bridging low-level episodic records and high-level semantic abstractions.

In contrast, Agent Semantic Memory stores \textit{crystallized knowledge} in two complementary forms: \textit{reusable response strategies} and \textit{topic-level user personas}.
Formally, we define the reusable response strategy under topic $t$ as
\begin{equation}
\begin{aligned}
\operatorname{Strategy}_{t} &= \bigl\{\langle t_j, i_j \rangle \mid {} \\
    &\quad \text{EM}_{\text{agent}}^{(j)} \in \text{EM}_{\text{agent}},\ t_j = t\bigr\}.
\end{aligned}
\end{equation}
where $t_j$ and $i_j$ denote the topic and insight fields of $\text{EM}_{\text{agent}}^{(j)}$, respectively. This definition retains only the topic-specific insight signals and removes the full interaction trajectory.

Similarly, the topic-level user persona is defined as
\begin{equation}
\operatorname{Persona}_{t}
=
\left\{
s_j
\mid
s_j \in \text{SM}_{\text{user}},
\ t_j = t
\right\},
\end{equation}
where $t_j$ denotes the topic associated with the user semantic memory $s_j$. Thus, the persona is represented directly as the set of user semantic memories under topic $t$, without an additional aggregation operator.

Accordingly, the agent semantic memory can be represented as
\begin{equation}
\text{SM}_{\text{agent}}^{(t)}
=
\langle
\operatorname{Strategy}_{t},
\operatorname{Persona}_{t}
\rangle .
\end{equation}

The strategy component encapsulates interaction patterns induced from historical episodes, while the persona component aggregates stable user preferences under specific topics.
By maintaining only these high-level abstractions, semantic memory minimizes redundancy and enables the agent to efficiently retrieve lightweight, executable instructions during inference.
This design effectively balances the expressiveness of detailed historical records with the execution efficiency of strategic reasoning.

\subsubsection{Memory Retrieval}
\label{app:memory_retrieval}

Figure~\ref{fig:Workflow} illustrates the overall memory retrieval process during inference.
Given a user query, the system first retrieves relevant information from agent memory to obtain strategy-level guidance, and then performs parallel retrieval over the user memory graph through multiple retrieval actions.
This retrieval order allows the agent to leverage accumulated interaction patterns before interacting with user-specific memories.

The retrieved agent memory provides high-level strategy signals that influence both the selection of retrieval targets and the generation of responses.
As shown in the workflow, different strategies may lead the agent to attend to different subsets of user memories and emphasize different aspects of retrieved information.
In this way, the execution workflow explicitly supports strategy-conditioned memory retrieval.

To support flexible strategy execution, the agent interacts with the memory system through a set of structured tools.
These tools enable the agent to perform specialized retrieval operations, such as searching for semantically related memories, identifying causally related episodic events, and traversing parent--child relations among episodic nodes.
By composing these tools across multiple retrieval steps, the agent can adaptively explore the memory graph and retrieve information most relevant to the current query and inferred strategy.

\subsubsection{Memory Evolution}
\label{app:memory_evolution}
Memory evolution in PRISM consists of interaction pool evolution and model evolution.
While interaction pool evolution operates at the memory level, model evolution addresses the personalization requirements of on-device large language models by incorporating long-term interaction signals into the model update process.

\paratitle{Interaction Pool Evolution.}

The goal of interaction pool evolution is to maintain a compact, high-quality, and learnable experience repository over long-term interactions.
As user interactions continuously accumulate, simply retaining all historical trajectories may introduce redundant, outdated, or low-value memory units.
Such uncontrolled accumulation can reduce retrieval efficiency and may further interfere with strategy induction and model updates.
Therefore, PRISM dynamically evaluates each agent episodic memory in the interaction pool and determines whether it should be retained according to a composite quality score.

For each agent episodic memory unit in the interaction pool, we compute a three-dimensional value vector:
\begin{equation}
V_{\text{EM}} = \langle T, I, L \rangle,
\end{equation}
where $T$ denotes temporal utility, $I$ denotes importance, and $L$ denotes learnability.
These three dimensions evaluate whether a memory should be retained from the perspectives of temporal relevance, informational value, and learning benefit, respectively.

\paratitle{Temporal Utility.}
Temporal utility $T$ jointly considers the recency and access frequency of a memory.
The recency score measures how recent a memory is and is defined as:
\begin{equation}
T_{\text{recency}} = \frac{1}{1 + \log(1 + \Delta t)},
\end{equation}
where $\Delta t$ denotes the elapsed time in days since the memory was created.
This function gradually decays as time increases, assigning higher scores to more recent memories while avoiding excessive penalization of older memories that may still remain useful.

The access frequency score measures how frequently a memory has been retrieved or used during previous reasoning processes:
\begin{equation}
T_{\text{frequency}} = \frac{\text{access\_count}}{\text{max\_access}},
\end{equation}
where $\text{access\_count}$ denotes the number of times the memory has been accessed, and $\text{max\_access}$ denotes the maximum access count within the current batch.

The final temporal utility score is computed by combining normalized recency and frequency scores:
\begin{equation}
T =
\frac{
\mathcal{N}(T_{\text{recency}})
+
\mathcal{N}(T_{\text{frequency}})
}{2},
\end{equation}
where $\mathcal{N}(\cdot)$ denotes min-max normalization within the current batch.

\paratitle{Importance.}
Importance $I$ measures whether a memory contains distinctive and valuable information that is worth preserving over the long term.
It consists of two components: information uniqueness and explicit importance.

Information uniqueness is computed based on the maximum semantic similarity between the current memory and other memories:
\begin{equation}
I_{\text{uniqueness}}
=
1 - \max_{j \neq i} \cos(e_i, e_j),
\end{equation}
where $e_i$ and $e_j$ denote the embedding representations of the $i$-th and $j$-th memories, respectively.
If a memory is highly similar to existing memories, its uniqueness score is lower; otherwise, a larger semantic distance indicates that the memory contains more distinctive information.

The explicit importance score is defined as:
\begin{equation}
I_{\text{importance}} \in [0, 1],
\end{equation}
which can be obtained from an LLM-based judge, user feedback, or explicit annotation signals.
This score estimates the direct importance of a memory for understanding user needs, user preferences, or future decision making.

The final importance score is computed as:
\begin{equation}
I =
\frac{
\mathcal{N}(I_{\text{uniqueness}})
+
\mathcal{N}(I_{\text{importance}})
}{2}.
\end{equation}

\paratitle{Learnability.}
Learnability $L$ measures whether reusing a memory can bring effective gains in subsequent tasks.
For each memory, the system records its usage count $\text{use\_count}$ and the number of successful outcomes after reuse, denoted as $\text{success\_count}$.
Learnability is defined as:
\begin{equation}
L =
\frac{\text{success\_count}}{\text{use\_count} + \varepsilon},
\end{equation}
where $\varepsilon$ is a small smoothing term used to avoid division by zero.
When $\text{use\_count}=0$, $L$ is set to $0.5$ by default, so that newly added memories that have not yet been sufficiently used are not prematurely penalized.

\paratitle{Composite Quality Score.}
After obtaining $T$, $I$, and $L$, PRISM converts the multi-dimensional value vector into a scalar quality score through a weighted aggregation function:
\begin{equation}
S_{\text{EM}}
=
w_T \cdot \mathcal{N}(T)
+
w_I \cdot \mathcal{N}(I)
+
w_L \cdot \mathcal{N}(L),
\end{equation}
where $w_T$, $w_I$, and $w_L$ denote the weights of temporal utility, importance, and learnability, respectively, subject to:
\begin{equation}
w_T + w_I + w_L = 1.
\end{equation}

By default, we set:
\begin{equation}
w_T = 0.35,\quad w_I = 0.40,\quad w_L = 0.25.
\end{equation}
The three dimensions are normalized again within the current batch before aggregation, ensuring numerical comparability across dimensions and preventing any single dimension from dominating the final score due to scale differences.

\paratitle{Retention and Removal Rule.}
Based on the composite quality score $S_{\text{EM}}$, PRISM applies a threshold-based filtering mechanism to update the interaction pool.
For an arbitrary agent episodic memory, it is retained if its quality score satisfies:
\begin{equation}
S_{\text{EM}} \geq \alpha.
\end{equation}
Otherwise, the memory is removed from the interaction pool.
By default, the threshold is set to:
\begin{equation}
\alpha = 0.50.
\end{equation}

Through this mechanism, the interaction pool can continuously receive new interactions while gradually filtering out low-value, redundant, or ineffective historical trajectories.
As a result, the memory repository remains compact and learnable.
Overall, interaction pool evolution functions not only as a memory compression mechanism, but also as an experience selection mechanism for long-term personalized learning.

\paratitle{Model Evolution.}
Model evolution in PRISM adapts the on-device language model to individual users using Group Relative Policy Optimization (GRPO).
Let the on-device model be parameterized by $\theta$, denoted as $\pi_\theta$.
Model updates are driven by agent episodic memories sampled from the evolved interaction pool.

For an agent episodic memory unit $\mathrm{EM}_j$ sampled from the interaction pool, the model generates a group of $G$ candidate trajectories
\begin{equation}
\{\tau_j^{(g)}\}_{g=1}^{G},
\end{equation}
each conditioned on the user query $Q_j$.
Each trajectory is evaluated using a personalization-aware reward function that incorporates user feedback $F_j$ and the derived insight signal $i_j$.

Under the GRPO framework, model parameters are updated according to:
\begin{equation}
\begin{aligned}
\theta \leftarrow {}& \theta + \eta \,
    \mathbb{E}_{\mathrm{EM}_j \sim p(\mathcal{P})}\biggl[ \\
    &\quad \frac{1}{G}\sum_{g=1}^{G}\tilde{A}_j^{(g)}\nabla_\theta
    \log \pi_\theta\!\left(\tau_j^{(g)} \mid Q_j\right)\biggr],
\end{aligned}
\end{equation}
where $\eta$ denotes the learning rate and $\tilde{A}_j^{(g)}$ is the group-relative advantage term.

The relative advantage is computed by normalizing rewards within each group:
\begin{equation}
\tilde{A}_j^{(g)}
=
R\!\left(\tau_j^{(g)}; F_j, i_j \right)
-
\frac{1}{G}
\sum_{g'=1}^{G}
R\!\left(\tau_j^{(g')}; F_j, i_j \right),
\end{equation}
where $R(\cdot)$ denotes a personalization-aware reward function.
By using user feedback on the original response and the derived insight signal as part of the reward signal, the model progressively internalizes long-term user preferences, which is particularly important in on-device scenarios.

\subsection{Core Algorithms}
\label{app:algorithms}

In this section, we present the pseudocode of the core algorithms underlying the PRISM framework.
Specifically, PRISM consists of five key procedures:
(1) user memory graph construction (Algorithm~\ref{alg:user_memory}),
(2) agent memory construction and abstraction (Algorithm~\ref{alg:agent_memory}),
(3) retrieval-augmented reasoning and execution (Algorithm~\ref{alg:retrieval}),
(4) interaction pool evolution (Algorithm~\ref{alg:pool_evolution}),
and (5) model evolution via GRPO-based reinforcement learning (Algorithm~\ref{alg:model_evolution}).
Together, these algorithms describe the full lifecycle of memory construction, reasoning, and self-evolution in PRISM.

\renewcommand{\algorithmicrequire}{\textbf{Input:}}
\renewcommand{\algorithmicensure}{\textbf{Output:}}
\newcommand{\FuncCall}[1]{\textsc{#1}}
\algrenewcommand{\algorithmiccomment}[1]{\hfill \textcolor{gray}{$\rightarrow$ #1}}

\begin{algorithm}[t]
\caption{PRISM System: User Memory Graph Construction}
\label{alg:user_memory}
\begin{algorithmic}[1]
\Require Raw interaction text $T$, user memory graph $G_{user}$
\Ensure Updated user memory graph $G_{user}$

\Procedure{BuildUserMemory}{$T$}
    \State Split $T$ into chunks $\mathcal{C}$
    \For{each chunk $c \in \mathcal{C}$}
        \State Extract topic $t$, semantic memory $S$, and episodic memory $E$ from $c$
        \State Update or create topic node $t$ in $G_{user}$

        \State Determine and execute the graph update for semantic memory $S$ under topic $t$
        \State Determine and execute the graph update for episodic memory $E$ under topic $t$
    \EndFor
    \State \Return $G_{user}$
\EndProcedure
\end{algorithmic}
\end{algorithm}

\begin{algorithm}[t]
\caption{PRISM System: Agent Memory Construction and Abstraction}
\label{alg:agent_memory}
\begin{algorithmic}[1]
\Require Raw interaction text $T$, agent episodic store $\mathcal{V}_{\mathrm{agent}}$, user memory graph $\mathcal{G}_{\mathrm{user}}$
\Ensure Updated $\mathcal{V}_{\mathrm{agent}}$ and agent semantic memory $\mathcal{S}_{\mathrm{agent}}$

\Procedure{BuildAgentMemory}{$T$}
    \State Split $T$ into chunks $\mathcal{C}$
    \For{each chunk $c \in \mathcal{C}$}
        \State Parse $c$ into user query $Q$, reasoning/retrieval trace $S$, agent response $A$, user feedback $F$, and timestamp $\tau$
        \State Derive a topic $t$ and an insight signal $i$ from $(Q, S, A, F, \tau)$
        \State Form an episodic memory record $\langle Q, S, A, F, t, i\rangle$
        \State Store the record into agent episodic store $\mathcal{V}_{\mathrm{agent}}$
    \EndFor
    \State Retrieve user semantic memories from $\mathcal{G}_{\mathrm{user}}$
    \State Aggregate user semantics into a persona representation
    \State Retrieve agent episodic memories from $\mathcal{V}_{\mathrm{agent}}$
    \State Induce a high-level strategy from episodic trajectories
    \State $\mathcal{S}_{\mathrm{agent}} \leftarrow \{\mathrm{persona}, \mathrm{strategy}\}$
    \State \Return $\mathcal{V}_{\mathrm{agent}}, \mathcal{S}_{\mathrm{agent}}$
\EndProcedure
\end{algorithmic}
\end{algorithm}

\begin{algorithm}[t]
\caption{Retrieval-Augmented Parallel Reasoning Process}
\label{alg:retrieval}
\begin{algorithmic}[1]
\Require User query $q$, agent semantic memory $\mathcal{S}_{\mathrm{agent}}$, agent episodic memory $\mathcal{E}_{\mathrm{agent}}$, user memory graph $\mathcal{G}_{\mathrm{user}}$, retrieval tool set $\mathcal{T}$, number of parallel branches $K$
\Ensure Final answer $A$

\Procedure{ReasonAndRetrieve}{$q, \mathcal{S}_{\mathrm{agent}}, \mathcal{E}_{\mathrm{agent}}, \mathcal{G}_{\mathrm{user}}, \mathcal{T}, K$}

    \State $\mathcal{R}^{\mathrm{ASM}} \leftarrow {}$
    \Statex \hspace{\algorithmicindent}\textsc{RetrieveAgentSemanticMemory}
    \Statex \hspace{\dimexpr\algorithmicindent*2\relax}$(\mathcal{S}_{\mathrm{agent}}, q)$
    \State $\mathcal{R}^{\mathrm{AEM}} \leftarrow {}$
    \Statex \hspace{\algorithmicindent}\textsc{RetrieveAgentEpisodicMemory}
    \Statex \hspace{\dimexpr\algorithmicindent*2\relax}$(\mathcal{E}_{\mathrm{agent}}, q)$
    \State $C_{\mathrm{agent}} \leftarrow \{\mathcal{R}^{\mathrm{ASM}}, \mathcal{R}^{\mathrm{AEM}}\}$

    \State $\{\alpha_k\}_{k=1}^{K} \leftarrow {}$
    \Statex \hspace{\algorithmicindent}\textsc{GenerateRetrievalActions}
    \Statex \hspace{\dimexpr\algorithmicindent*2\relax}$(q, C_{\mathrm{agent}}, \mathcal{T})$

    \For{$k = 1$ to $K$ \textbf{ in parallel}}
        \State $\mathcal{R}^{\mathrm{UM}}_{k} \leftarrow {}$
        \Statex \hspace{\dimexpr\algorithmicindent*2\relax}\textsc{ExecuteGraphRetrieval}
        \Statex \hspace{\dimexpr\algorithmicindent*3\relax}$(\alpha_k, \mathcal{G}_{\mathrm{user}})$
    \EndFor

    \State $\mathcal{R}^{\mathrm{UM}} \leftarrow \bigcup_{k=1}^{K} \mathcal{R}^{\mathrm{UM}}_{k}$

    \State $A \leftarrow \textsc{GenerateAnswer}$
    \Statex \hspace{\dimexpr\algorithmicindent*2\relax}$(q, \mathcal{R}^{\mathrm{ASM}}, \mathcal{R}^{\mathrm{AEM}}, \mathcal{R}^{\mathrm{UM}})$

    \State \Return $A$
\EndProcedure
\end{algorithmic}
\end{algorithm}

\begin{algorithm}[t]
\caption{PRISM System: Interaction Pool Evolution}
\label{alg:pool_evolution}
\begin{algorithmic}[1]
\Require Interaction Pool $\mathcal{P}$, Quality Threshold $\tau$
\Ensure Updated Interaction Pool $\mathcal{P}'$

\Procedure{\FuncCall{EvolveInteractionPool}}{$\mathcal{P}, \tau$}
    \For{each episodic memory $\mathrm{EM}_j \in \mathcal{P}$}
        \State $T_j \leftarrow$ temporal utility of $\mathrm{EM}_j$
        \State $I_j \leftarrow$ importance score of $\mathrm{EM}_j$
        \State $L_j \leftarrow$ learnability score of $\mathrm{EM}_j$
        \State $S_{\mathrm{EM}_j} \leftarrow \phi(T_j, I_j, L_j)$
        \If{$S_{\mathrm{EM}_j} < \tau$}
            \State $\mathcal{P} \leftarrow \mathcal{P} \setminus \{\mathrm{EM}_j\}$
        \EndIf
    \EndFor
    \State \Return $\mathcal{P}$
\EndProcedure
\end{algorithmic}
\end{algorithm}

\begin{algorithm}[t]
\caption{PRISM System: Model Evolution via GRPO}
\label{alg:model_evolution}
\begin{algorithmic}[1]
\Require Interaction Pool $\mathcal{P}$, Policy $\pi_\theta$, Group Size $G$
\Ensure Updated Policy $\pi_{\theta^*}$

\Procedure{\FuncCall{EvolveModel}}{$\mathcal{P}, \pi_\theta$}
    \While{training not converged}
        \State Sample episodic memory $\mathrm{EM}_j \sim p(\mathcal{P})$
        \State Extract query $Q_j$, feedback $F_j$, insight $i_j$ from $\mathrm{EM}_j$
        \State Generate $G$ candidate trajectories $\{\tau_j^{(g)}\}_{g=1}^{G}$ using $\pi_\theta$
        \For{each trajectory $\tau_j^{(g)}$}
            \State Compute reward $R_j^{(g)} \leftarrow R(\tau_j^{(g)}; F_j, i_j)$
        \EndFor
        \State Update policy $\pi_\theta$ using GRPO
    \EndWhile
    \State \Return $\pi_\theta$
\EndProcedure
\end{algorithmic}
\end{algorithm}

\subsection{Additional Case Studies}   
This case, observed on the Mem-PAL dataset and illustrated in Figure~\ref{fig:appendix_case1}, demonstrates the important role of agent memory in guiding memory retrieval through learned strategies, while also highlighting the critical value of multi-perspective memory retrieval in producing comprehensive responses.
By retrieving relevant interaction trajectories, PRISM develops a holistic understanding of the user’s needs, particularly in recognizing the pressure that work places on family life and personal health, as well as the need for effective time management.
Consequently, PRISM achieves a high level of personalization, whereas Mem0 produces a more generic response.

\label{app:case_study}
\begin{figure*}[t]
\centering
\begin{tcolorbox}[
    colback=white, 
    colframe=black!75, 
    fonttitle=\bfseries,
    title={Case Study: Requirement Restatement(mempal-id:45)},
    arc=0mm,
    boxrule=0.8pt,
    width=1.0\textwidth,
    left=2mm, right=2mm, top=2mm, bottom=2mm
]
    \small
    \textbf{User Query:} \textit{``I’ve been feeling that I never manage to finish my work plans. Is there a better way to organize my time?''} \\
    \textbf{Golden Answer:}
    The user hopes to learn more systematic task decomposition and time planning methods in order to complete work efficiently while reducing psychological burden, thereby achieving a better work–life balance.\\
    \textbf{Evidence\&Analysis:}
    The user previously reported low productivity and difficulty initiating large tasks. Moreover, work-related pressures have interfered with family life and adversely impacted personal health. As a result, the user seeks more effective task management strategies and improved work–life balance.

    \vspace{2mm}
    \begin{tabularx}{\linewidth}{X | X}
    \rowcolor{gray!10}
    \textbf{Mem0} & \textbf{Prism} \\
    \midrule
    
    \textbf{Retrieved Memory:} &
    \textbf{Retrieved Memory:} \\[0.5em]
    
    $\bullet$ \textit{I’ve been feeling that I can’t finish my work plans lately.} 
    &
    $\bullet$ \textbf{Agent Memory:} 
    \textit{Effective recommendations should address not only time efficiency but also the user’s health and family-related concerns.} \\[0.5em]
    
    $\bullet$ \textit{Recently, I’ve been struggling to balance family activities and work.} 
    &
    $\bullet$ \textbf{Agent Memory:} 
    \textit{Ignoring the user’s need for family time may lead to solutions that do not align with user expectations.} \\[0.5em]
    
    $\bullet$ \textit{I hope to find concrete methods to improve efficiency so that I can spend more time with my family.} 
    &
    $\bullet$ \textbf{Semantic Memory:} 
    \textit{The user requires time management methods that fit their personal work rhythm.} \\[0.5em]
    
    $\bullet$ \textit{I hope to manage my time better in the new year.} 
    &
    $\bullet$ \textbf{Semantic Memory:} 
    \textit{The user is actively attempting to adjust their time allocation.} \\[0.5em]
    
    $\bullet$ \textit{I like setting aside a dedicated ``family golden time'' to minimize work-related interruptions.}
    &
    $\bullet$ \textbf{Event Memory:} 
    \textit{The user has recently felt short on time and is seeking a more reasonable schedule that allows both work completion and sufficient time for family interaction, while also considering healthier daily routines.} \\
    
    \addlinespace[1mm]
    \midrule
        \addlinespace[1mm]
        {\color{red!70!black} \textbf{Model Response (1/2):}} & {\color{green!40!black} \textbf{Model Response (2/2):}} \\
        \textit{``The user hopes to alleviate conflicts between work and family life and reduce work-related stress through effective strategies.''} & \textit{``The user seeks time management methods aligned with their personal work rhythm, aiming to establish clear boundaries for family time and personal health without increasing pressure, thereby improving work–family balance and overall well-being.''} \\
    \end{tabularx}
\end{tcolorbox}
\vspace{-8pt}
\caption{An example of requirement restatement for a work-related query.}
\label{fig:appendix_case1}
\end{figure*}

\section{Experimental Details and Supplementary Experiments}
\label{app:experiment}

\subsection{Dataset and Baseline Details}
\label{app:dataset}
\paratitle{LangMem.}
LangMem \cite{langmem2024github} is a long-term memory module in LangChain that extracts and stores key information from conversations for later retrieval. It provides storage-agnostic functional primitives and integrates natively with LangGraph's storage layer. In our experiments, we use Qdrant to persist memories.

\paratitle{Mem0.}
Mem0 \cite{mem0} is a memory-centric framework that extracts, organizes, and retrieves user-related information to support long-term personalization in LLM-based agents. In our experiments, we use the entity-based version of Mem0 as a graph-based memory construction baseline, where user-related entities and relations are organized into a structured memory graph.

\paratitle{A-MEM.}
A-MEM \cite{xu2025mem} is an agentic memory framework for LLM agents that dynamically constructs, organizes, and retrieves memories. Instead of relying only on static memory operations, A-MEM introduces an adaptive memory organization mechanism to associate related memory units and support more flexible memory retrieval. In our experiments, we include A-MEM as a representative agentic memory baseline and evaluate it under the same backbone settings as other memory-based methods.

\paratitle{Mem-$\alpha$.}
Mem-$\alpha$ \cite{wang2025mem} is a reinforcement-learning-based memory framework that trains agents to construct and manage memories according to downstream task feedback. Since Mem-$\alpha$ requires reinforcement learning for optimization, we evaluate it only under the Qwen2.5-7B-Instruct setting, which is consistent with the local backbone setting used for our RL experiments.

\paratitle{Base+RAG.}
To enable Qwen2.5-7B-Instruct to complete basic tasks, we employ a Retrieval-Augmented Generation (RAG) approach. We store each dialogue turn or information unit and retrieve relevant entries by calculating the cosine similarity between the stored items and the current query, following standard RAG practices.

\paratitle{Mem-PAL.}
Mem-PAL \cite{huang2025mem} primarily evaluates an agent's ability to memorize, update, and utilize user preferences and user-related information over long-term interactions. In our experiments, we adopt the Requirement Restatement and Solution Proposal tasks, and utilize LLM-as-a-Judge to perform quantitative automatic evaluation on the Multi-turn Dialogue Interaction setting.

\paratitle{KnowMeBench.}
KnowMeBench \cite{wu2026knowme} is a benchmark designed to evaluate an agent's personalized reasoning capability by organizing questions into three hierarchical levels:
(1) Accuracy and Factuality, which assesses precise memory retrieval and resistance to hallucination via tasks like Context-aware Extraction, Adversarial Abstention, and Temporal Reasoning;
(2) Narrative Logic and Causality, which examines the understanding of event progression and associative connections through Logical Event Ordering and Memory Trigger Analysis; and
(3) Psychoanalytic Depth, which tests high-level interpretation of implicit motives and psychological states using Somatic-Psychic Interaction and Expert-Annotated Psychoanalysis.
By structuring queries across these levels, KnowMeBench enables the evaluation of both answer correctness and the agent's ability to produce coherent, user-aware reasoning and insights. Given the open-ended nature of most questions, we employ LLM-as-a-Judge to perform quantitative automatic evaluation.

\paratitle{LoCoMo.}
LoCoMo \cite{maharana2024evaluating} is a benchmark for evaluating very long-term conversational memory in LLM agents. It contains long conversational histories and questions that require models to retrieve, integrate, and reason over information from extended interactions. In our supplementary experiments, we evaluate model performance across four question types, including single-hop, multi-hop, open-domain, and temporal questions, and report LLM-as-a-Judge accuracy as the main metric.
\subsection{Experimental Setup Details}
\label{app:setup-details}

\begin{table}[t]
\centering
\small
\setlength{\tabcolsep}{2.5pt}
\begin{tabular}{ll}
\toprule
Tool / Library / Hyperparameter & Value \\
\midrule
Neo4j & 5.21 \\
verl & 6.0.0 \\
qdrant-client & 1.9 \\
PyTorch & 2.7 \\
Number of parallel retrieval branches $K$ & 5 \\
TIL weights ($w_T$ / $w_I$ / $w_L$) & 0.35 / 0.40 / 0.25 \\
Evolution threshold $\alpha$ & 0.50 \\
\bottomrule
\end{tabular}
\caption{Implementation details and main hyperparameter settings.}
\label{tab:implementation_details}
\end{table}

Table~\ref{tab:implementation_details} summarizes the main tools, libraries, and default hyperparameter settings used in our implementation. 
We use Neo4j for structured user memory storage, Qdrant for vector-based agent memory retrieval, and verl for reinforcement learning training. 
The default TIL weights and evolution threshold follow the interaction pool evolution design described in Appendix~\ref{app:memory_evolution}.

\paratitle{Backbone-specific Experimental Settings.}
To avoid ambiguity between inference-time memory comparison and RL-based model evolution, we distinguish the roles of different backbone settings.
GPT-5-mini is used as a fixed inference-time backbone to evaluate the effectiveness of different memory mechanisms, and we do not perform policy-gradient updates on this API-based model.
In contrast, Qwen2.5-7B-Instruct serves as the local trainable policy model for reinforcement learning experiments.
Therefore, RL-based methods, including Mem-$\alpha$ and PRISMRL, are evaluated only under the Qwen2.5-7B-Instruct setting.
For the Qwen setting, Base+RAG denotes a retrieval-augmented non-memory baseline that provides basic task-solving ability under the local backbone.
Within each backbone setting, all methods are evaluated using the same input queries, memory sources, and benchmark-specific evaluation protocols.

\paratitle{Reinforcement Learning Implementation.}
For reinforcement learning, we use the verl framework and choose Qwen2.5-7B-Instruct as the backbone policy model.
The model is trained on 8 NVIDIA H100 GPUs with a learning rate of $1\mathrm{e}{-6}$, a batch size of 32, and a GRPO rollout number of 8.
The complete training process consists of 200 steps, and we select the best checkpoint according to validation performance.
The training data is constructed from the history training split of Mem-PAL.

\subsection{Experimental Evaluation Details}
\label{app:evaluation-details}
\begin{table}[t]
\centering
\caption{Agreement between LLM-as-a-Judge and human annotation.}
\label{tab:judge-human-agreement}
\renewcommand{\arraystretch}{1.12}
\setlength{\tabcolsep}{3.5pt}
\resizebox{\linewidth}{!}{%
\begin{tabular}{lcccc}
\toprule
\textbf{Benchmark} & \textbf{Scale} & \textbf{Samples} & \textbf{Agreement Rate (\%)} & \textbf{Mean Absolute Deviation} \\
\midrule
Mem-PAL       & 0--2 & 200 & 95.0 & 0.06 \\
KnowMeBench   & 0--5 & 200 & 92.0 & 0.09 \\
\bottomrule
\end{tabular}%
}
\vspace{-5mm}
\end{table}

\paragraph{LLM-based Evaluation Protocol.}
For open-ended and preference-sensitive tasks, lexical-overlap metrics alone are often insufficient to capture semantic correctness and personalized response quality. 
Therefore, we adopt LLM-as-a-Judge evaluation following the official evaluation prompts and scoring rubrics provided by the corresponding benchmarks. 
All compared methods are evaluated under the same benchmark-specific judge prompt, ensuring that the automatic evaluator is applied consistently across methods.

During benchmark evaluation, the judge only receives the user context, the user query, the model's final answer, and the reference answer when available. 
Hidden chain-of-thought, intermediate retrieval traces, memory retrieval results, and method-specific internal states are not exposed to the judge. 
This design prevents the evaluator from accessing information that is unavailable in the final response and avoids giving memory-based methods additional evaluation-side advantages.

\paragraph{Separation between Training Reward and Benchmark Evaluation.}
Although both the RL stage and the final evaluation involve LLM-based scoring, they serve different purposes and are kept strictly separated.
During RL training, the persona reward is used only as a training-time signal to evaluate candidate trajectories sampled from the local policy model, with the goal of encouraging consistency with user feedback and inferred user preferences.
During benchmark evaluation, we follow the official prompts and scoring rubrics of each benchmark.
The evaluation judge does not receive the RL reward prompt, sampled trajectories, training-time feedback signals, intermediate retrieval traces, hidden reasoning states, or method-specific memory states.
Thus, benchmark scores are not computed by reusing the RL reward prompt, which reduces the risk of reward--evaluator coupling.

\paragraph{Human Validation.}
To further examine the reliability of the automatic evaluation, we conduct a human validation study on randomly sampled examples. 
Specifically, we sample 200 examples from Mem-PAL and 200 examples from KnowMeBench, and ask human annotators to assess both answer correctness and personalization consistency based on the same user context and reference information. 
As shown in Table~\ref{tab:judge-human-agreement}, the LLM-as-a-Judge results show high agreement with human annotations, with agreement rates of 95.0\% on Mem-PAL and 92.0\% on KnowMeBench. 
The mean absolute deviations are also low, at 0.06 and 0.09, respectively. 
These results indicate that the automatic evaluation is generally consistent with human judgments, while also providing additional evidence that the gains of PRISM are not merely due to stylistic alignment.

\subsection{Supplementary Experiments}
\label{app:supplementary}

To further analyze the generalization ability, practical applicability, and robustness of the proposed method under different settings, we conduct a series of supplementary experiments. Specifically, these experiments cover four aspects:
(1) the performance on the LoCoMo dataset \cite{maharana2024evaluating};
(2) the time and token overhead of PRISM on the LoCoMo dataset;
(3) the effect of different reward hyperparameters; and
(4) the performance of our method on smaller-scale backbone models.
Together, these supplementary results provide a more comprehensive understanding of the effectiveness, efficiency, and scalability of our method beyond the main experimental setting.

\subsubsection{Performance on LoCoMo}
\label{app:locomo}

To further examine whether PRISM generalizes beyond the main evaluation datasets, we conduct additional experiments on LoCoMo \cite{maharana2024evaluating}, a benchmark for evaluating very long-term conversational memory in LLM agents. We report LLM-as-a-Judge accuracy across four question types, including single-hop, multi-hop, open-domain, and temporal questions.

As shown in Table~\ref{tab:locomo}, PRISM achieves the best performance across all question types under GPT-5-mini and obtains an overall accuracy of 78.96. Under Qwen2.5-7B-Instruct, PRISMRL also achieves the best performance across all question types, with an overall accuracy of 53.57. These results demonstrate that the proposed memory mechanism generalizes to long-context conversational memory tasks across both API-based and local backbone settings.

\begin{table}[t]
\centering
\caption{Performance comparison on LoCoMo. The \tc{bestcolor}{best} and \tc{secondcolor}{second-best} results are highlighted.}
\label{tab:locomo}
\renewcommand{\arraystretch}{1.05}
\setlength{\tabcolsep}{3.0pt}

\resizebox{\columnwidth}{!}{%
\begin{tabular}{llccccc}
\toprule
\multirow{2}{*}{\textbf{Backbone}} & \multirow{2}{*}{\textbf{Method}} & \multicolumn{5}{c}{\textbf{LoCoMo}} \\
\cmidrule(lr){3-7}
 & & \textbf{Single Hop} & \textbf{Multi-Hop} & \textbf{Open Domain} & \textbf{Temporal} & \textbf{Overall} \\
\midrule
\multirow{5}{*}{GPT-5-mini}
 & Base          & \cc{secondcolor}{86.80} & \cc{secondcolor}{68.44} & 55.21 & 50.47 & \cc{secondcolor}{73.90} \\
 & LangMem       & 62.07 & 47.87 & \cc{secondcolor}{70.83} & 23.36 & 51.95 \\
 & Mem0 (entity) & 66.23 & 56.74 & 44.79 & 58.88 & 61.62 \\
 & A-MEM         & 71.82 & 56.38 & 32.29 & \cc{secondcolor}{60.12} & 64.09 \\
 & PRISM         & \cc{bestcolor}{86.92} & \cc{bestcolor}{68.79} & \cc{bestcolor}{76.04} & \cc{bestcolor}{67.91} & \cc{bestcolor}{78.96} \\
\midrule
\multirow{7}{*}{Qwen2.5-7B-Instruct}
 & Base+RAG      & 26.75 & 28.72 & 39.58 & 2.80 & 22.92 \\
 & LangMem       & 26.99 & 30.85 & 26.04 & 5.92 & 23.25 \\
 & Mem0 (entity) & 29.01 & 30.85 & 44.79 & 3.43 & 25.00 \\
 & A-MEM         & 45.18 & 46.45 & 52.08 & 9.35 & 38.38 \\
 & Mem-$\alpha$(RL) & 49.23 & 49.65 & 55.21 & 12.46 & 42.01 \\
 & PRISM         & \cc{secondcolor}{55.05} & \cc{secondcolor}{54.96} & \cc{secondcolor}{61.46} & \cc{secondcolor}{18.07} & \cc{secondcolor}{47.73} \\
 & PRISMRL       & \cc{bestcolor}{60.05} & \cc{bestcolor}{61.35} & \cc{bestcolor}{68.75} & \cc{bestcolor}{25.23} & \cc{bestcolor}{53.57} \\
\bottomrule
\end{tabular}%
}
\vspace{-5mm}
\end{table}

\subsubsection{Time and Token Overhead on LoCoMo}
\label{app:locomo_overhead}

We further analyze the token consumption and runtime overhead of different memory-based methods on LoCoMo. Specifically, we report the number of tokens used for memory storage, the number of tokens consumed during question answering, and the total QA runtime.

As shown in Table~\ref{tab:locomo_overhead}, PRISM introduces moderate storage and inference overhead while achieving the best performance in Table~\ref{tab:locomo}. Compared with baselines that require larger storage or QA token consumption, PRISM maintains a more balanced trade-off between effectiveness and efficiency.

\begin{table}[t]
\centering
\caption{Token and runtime overhead analysis on LoCoMo.}
\label{tab:locomo_overhead}
\renewcommand{\arraystretch}{1.05}
\setlength{\tabcolsep}{3.0pt}

\resizebox{\columnwidth}{!}{%
\begin{tabular}{lccc}
\toprule
\textbf{Method} & \textbf{Store Tokens (k)} & \textbf{QA Tokens (k)} & \textbf{Runtime (s)} \\
\midrule
LangMem          & 3,842.156  & 4,936.284  & 2,214 \\
Mem0             & 24,304.872 & 1,488.618  & 1,907 \\
A-MEM            & 11,494.344 & 10,170.567 & 6,477 \\
Mem-$\alpha$(RL) & 5,736.491  & 2,684.735  & 3,126 \\
PRISM            & 4,218.939  & 1,931.688  & 2,708 \\
\bottomrule
\end{tabular}%
}
\vspace{-5mm}
\end{table}

\subsubsection{Reward Hyperparameter Analysis}
\label{app:reward_hyper}

In our framework, the overall reward function consists of two components: a task accuracy reward and a user persona reward. To investigate the trade-off between these two objectives, we systematically vary their relative weights and evaluate the resulting performance on the KnowMeBench dataset.

\begin{figure}[t]
    \centering
    \resizebox{1.0\columnwidth}{!}{\includegraphics{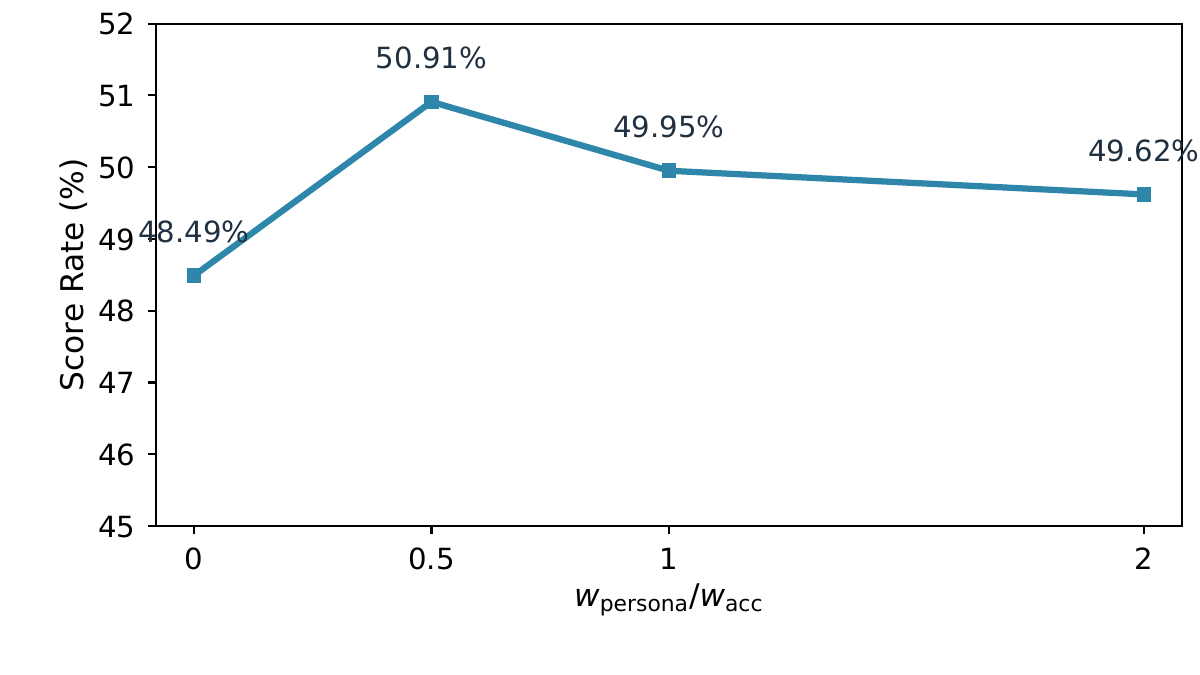}}
    \vspace{-6mm}
    \caption{Effect of different weighting ratios between accuracy reward and user persona reward on model performance.} 
    \label{fig:reward_ratio}
    \vspace{-4mm}
\end{figure}

As shown in Fig.~\ref{fig:reward_ratio}, we compare multiple configurations, including a setting that relies solely on the task accuracy reward and several settings that incorporate the user persona reward with different weight ratios. The results show that the model achieves the best overall performance when the task accuracy reward and the user persona reward are weighted with a ratio of $2{:}1$. In contrast, the configuration using only the task accuracy reward consistently underperforms those that additionally incorporate user persona information.

This observation suggests that explicitly introducing a user persona reward remains beneficial, even in task settings where task accuracy is the primary objective. By modeling user profiles, the agent is encouraged to favor solutions that better align with long-term user preferences among multiple plausible options, rather than optimizing solely for immediate correctness. As a result, user-aware reward signals can positively influence the agent’s reasoning and decision-making process without significantly compromising task accuracy.

Based on empirical observations on the KnowMeBench dataset, we adopt a $2{:}1$ ratio between the task accuracy reward and the user persona reward as the default setting, which provides a favorable balance between performance and training stability.

Beyond the reward weighting ratio, we further examine two hyperparameters used in interaction pool evolution: the TIL aggregation weights and the evolution threshold $\alpha$. 
As shown in Table~\ref{tab:til_weight_sensitivity}, the default TIL configuration achieves the best performance among the tested weight settings, indicating that balancing temporal utility, importance, and learnability is more effective than emphasizing any single dimension. 
Table~\ref{tab:evolution_threshold_sensitivity} further shows that $\alpha=0.50$ provides the best empirical performance, while either retaining too many low-value trajectories or removing too many potentially useful trajectories leads to performance degradation.

\begin{table}[t]
\centering
\small
\begin{tabular}{lcccc}
\toprule
Config & $w_T$ & $w_I$ & $w_L$ & Acc \\
\midrule
Equal & 0.333 & 0.333 & 0.333 & 29.63 \\
$T$-heavy & 0.600 & 0.250 & 0.150 & 28.38 \\
$I$-heavy & 0.200 & 0.650 & 0.150 & 30.75 \\
$L$-heavy & 0.200 & 0.250 & 0.550 & 29.13 \\
Default (TIL) & 0.350 & 0.400 & 0.250 & \textbf{32.38} \\
\bottomrule
\end{tabular}
\caption{Sensitivity analysis of the TIL weights used in interaction pool evolution.}
\label{tab:til_weight_sensitivity}
\end{table}

\begin{table}[t]
\centering
\small
\begin{tabular}{cc}
\toprule
Evolution threshold $\alpha$ & Acc \\
\midrule
0.30 & 29.25 \\
0.40 & 31.13 \\
0.50 & \textbf{32.63} \\
0.60 & 31.38 \\
0.70 & 29.75 \\
\bottomrule
\end{tabular}
\caption{Sensitivity analysis of the interaction pool evolution threshold $\alpha$.}
\label{tab:evolution_threshold_sensitivity}
\end{table}

In practical deployment scenarios, these hyperparameters can be adjusted by users or developers according to application-specific requirements. 
For example, applications that emphasize recent user behavior may assign a larger temporal utility weight, while resource-constrained settings may adopt a higher evolution threshold to maintain a more compact interaction pool.

\begin{figure}[t]
    \centering
    \resizebox{1.0\columnwidth}{!}{\includegraphics{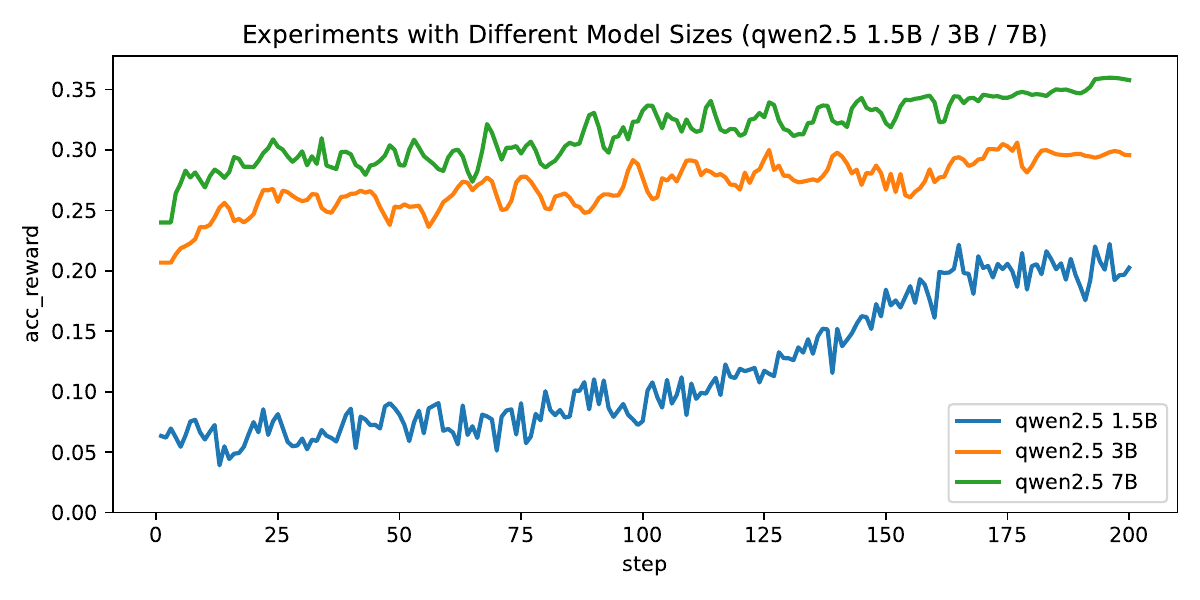}}
    \vspace{-6mm}
    \caption{Performance of the proposed method on smaller backbone models.} 
    \label{fig:small_models}
    \vspace{-4mm}
\end{figure}

\subsubsection{Experiments with Smaller Models}
\label{app:small_models}

In addition to the main experiments conducted with larger backbone models, we further evaluate our method on smaller-scale models with 1.5B and 3B parameters to assess its effectiveness under resource-constrained settings.

As illustrated in Fig.~\ref{fig:small_models}, our approach consistently improves performance on the Mem-PAL dataset for both the 1.5B and 3B models. These results demonstrate that the proposed method does not rely exclusively on the expressive capacity of large models, but remains effective even when applied to smaller backbones.

This finding indicates that smaller models are still capable of understanding and leveraging user persona information when guided by appropriate memory mechanisms and reward signals. With explicit user-aware supervision, even parameter-limited models can incorporate user preferences into their reasoning process, leading to improved task performance.

\section{Prompts}
\label{app:prompts}

In developing the prompts, we draw inspiration from the designs of Mem0~\cite{mem0} and SearchR1~\cite{jin2025search}.
For the LLM-as-a-Judge prompt used during the reinforcement learning training stage, we adapt the prompt released by the Mem-PAL dataset~\cite{huang2025mem}.
During benchmark evaluation, we directly adopt the official prompts and scoring rubrics provided by the corresponding datasets. Therefore, the training-time judge prompt is not reused as the benchmark evaluation prompt.
In this section, we present the following specific implementations:
(i) the Semantic Memory Extraction prompt (Figure~\ref{fig:semantic_prompt}),
(ii) the ReAct-paradigm reasoning prompt for personalized agents (Figure~\ref{fig:agent_prompt}),
and (iii) the training-time Judge prompt used to compute the persona reward during reinforcement learning (Figure~\ref{fig:judge_prompt}).
For other prompt implementations, please refer to our released repository.

\newtcolorbox{WidePrompt}[1]{
    enhanced,
    width=\textwidth,        
    colback=MidnightBlue!3,  
    colframe=MidnightBlue!30, 
    sharp corners,
    boxrule=0.5pt,
    fonttitle=\bfseries\sffamily,
    fontupper=\small\sffamily,
    title={\faTerminal\ #1},
    coltitle=MidnightBlue,    
    attach boxed title to top left={yshift=-2mm, xshift=3mm},
    boxed title style={
        colback=white,       
        colframe=MidnightBlue!30, 
        sharp corners,
        boxrule=0.5pt
    }
}

\begin{figure*}[t]
    \centering
    \begin{WidePrompt}{Extractor Prompt: Semantic Memory Extraction}
    You are an AI assistant specialized in extracting semantic knowledge from conversation blocks. The input consists of a list of dialogue entries. When generating semantic triples, for any fact involving the user, you MUST use the value of "\{user\}" as the subject. \\
    Definition of Semantic Memory: \\
    Semantic memory represents long-term facts only: \\
    1. Long-term personal preferences \\
    2. Long-term personal habits \\
    3. General world knowledge about external concepts \\
    If a statement does not explicitly imply stability or continuity, you MUST NOT extract it. \\
    Strict Exclusion Rules: \\
    The following content must NEVER be extracted as semantic facts: \\
    - One-time actions or single specific events \\
    - Actions tied to a specific time \\
    - Temporary emotions or physical states in a specific context \\
    - Future plans ("I will...", "I might...", "I am going to...") \\
    - Hypotheticals or conditional statements \\
    - Any content that is not clearly long-term and stable \\
    Format Requirement: \\
    Each semantic fact must follow this structure: [subject, predicate, object, topic] \\
    Subject: \\
    - If the fact describes the user, it MUST be the exact user string \\
    - The subject MAY be an external concept \\
    Predicate: \\
    - Must describe a stable relationship \\
    Object: \\
    - Must be a category, attribute, or stable concept \\
    Rules for topic: \\
    - The topic is used to summarize the category or domain of the fact, or a specific attribute of the user \\
    - It must contain one short label \\
    Example: ["Jack", "likes", "playing the piano", "leisure"] \\
    Input Block: \{block\_text\} \\
    Output Format (STRICT): \\
    (omitted for brevity)
    \end{WidePrompt}
    \caption{The prompt for extracting semantic memory from conversations.}
    \label{fig:semantic_prompt}
\end{figure*}


\newtcolorbox{WideJudge}[1]{
    enhanced,
    width=\textwidth,
    colback=RoyalPurple!2,       
    colframe=RoyalPurple!25,     
    sharp corners,
    boxrule=0.4pt,
    fonttitle=\bfseries\sffamily,
    fontupper=\small\sffamily,
    title={\faBalanceScale\ #1}, 
    coltitle=RoyalPurple!80,     
    attach boxed title to top left={yshift=-2mm, xshift=3mm},
    boxed title style={
        colback=white,
        colframe=RoyalPurple!25, 
        sharp corners,
        boxrule=0.4pt
    }
}

\newtcolorbox{AgentMemory}[2][]{
    enhanced,
    width=\textwidth,              
    colback=forestgreen!5,         
    colframe=forestgreen!40,       
    sharp corners,                 
    boxrule=0.4pt,                 
    fonttitle=\bfseries\sffamily,
    fontupper=\small\sffamily\linespread{1.05}\selectfont,     
    title={\faRobot\ \ #2},        
    coltitle=forestgreen,          
    attach boxed title to top left={yshift=-2mm, xshift=3mm}, 
    boxed title style={
        colback=white,             
        colframe=forestgreen!40,   
        sharp corners,
        boxrule=0.4pt
    },
    #1
}
\begin{figure*}[t]
    \centering
    \begin{AgentMemory}{Reasoning Prompt: React Agent}
    You are a personalized reasoning agent integrated with long-term memory retrieval capabilities. Your task is to combine user background with the current query, retrieve necessary information via tool calls, and provide a final response. \\
    1. Context Variables \\
    - Target User: \{username\} \\
    - Core Query: \{query\} \\
    - High-level Strategy: \{agentmemory\} \\
    - Status: Max steps \{max\_step\}, Current step \{current\_step\}. \\
    2. History of Observations \\
    \{observation\} \\
    \textit{--- The above are feedback from your previous actions; proceed with reasoning based on these results. ---} \\
    3. Execution Protocol \\
    - 1. Thought: \\
    - You must display your reasoning process within <think> tags. \\
    - Analyze Observations: Identify which pieces of information (preferences, habits, or background) are still missing. \\
    - Strategy Application: Refer to the suggestions in Strategy. If they can guide your current retrieval or judgment, adopt them; if they are irrelevant or unhelpful, feel free to ignore them. \\
    - 2. Action or Answer: \\
    - If the information currently acquired is insufficient to answer the query, you may call tools to retrieve more information. \\
    - When you believe the information is sufficient to address the user's request, or when the maximum step limit is reached, you must stop retrieving and provide the final answer. \\
    - The final answer must be wrapped in <answer> tags. \\
    \end{AgentMemory}
    \caption{The prompt for the Memory Agent.}
    \label{fig:agent_prompt}
\end{figure*}

\begin{figure*}[t]
    \centering
    \begin{WideJudge}{Training Reward Judge Prompt (LLM-as-a-Judge)}
    You are an evaluator. Please assign a score to the model response by considering the user's profile and preferences together with the question. \\
    User Profile and Preferences  \{persona\} \\
    Question \{question\} \\
    Model Reasoning \{think\} \\
    Model Final Answer \{answer\} \\
    Reference Answer \{ground\_truth\} \\
    Scoring Criteria \\
    - 0.5: The reasoning is consistent with the user's profile and preferences; the reasoning and the final answer are self-consistent; the answer matches the reference answer. \\
    - 0.3: The answer is correct, but shows some deviation from the user's profile and preferences. \\
    - 0.0: The answer is clearly inconsistent with the user's profile and preferences, or contains logical errors and an incorrect answer. \\
    Example: \\
    User profile: High work pressure and wants to spend more time with children \\
    Question: Looking for something to do during leisure time \\
    Reference answer: Choose badminton to exercise \\
    If the model chooses badminton but only discusses the benefits of badminton itself, without analyzing how exercise can relieve work stress or serve as a family activity, the score should be 0.3. \\
    Output Requirements \\
    (omitted for brevity)
    \end{WideJudge}
    \caption{The training-time LLM-as-a-Judge prompt used to compute the persona reward during reinforcement learning, rather than the benchmark evaluation prompt.}
    \label{fig:judge_prompt}
\end{figure*}

\end{document}